\documentclass{article}

\usepackage{microtype}
\usepackage{graphicx}
\usepackage{subfigure}
\usepackage{booktabs} 

\usepackage{hyperref}

\usepackage[accepted]{icml2024}

\usepackage{amsmath}
\usepackage{amssymb}
\usepackage{mathtools}
\usepackage{amsthm}

\usepackage[capitalize,noabbrev]{cleveref}

\theoremstyle{plain}

\theoremstyle{definition}

\theoremstyle{remark}

\usepackage[textsize=tiny]{todonotes}

\icmltitlerunning{Diverse Feature Learning by Self-distillation and Reset}

\begin{document}

\twocolumn[
\icmltitle{Diverse Feature Learning by Self-distillation and Reset}



\icmlsetsymbol{equal}{*}

\begin{icmlauthorlist}
\icmlauthor{Sejik Park}{sch}
\end{icmlauthorlist}

\icmlaffiliation{sch}{Graduate School of AI, Korea Advanced Institute of Science and Technology (KAIST), Seoul, South Korea}

\icmlcorrespondingauthor{Sejik Park}{sejik.park@kaist.ac.kr}

\icmlkeywords{Machine Learning, ICML}

\vskip 0.3in
]



\printAffiliationsAndNotice{}  

\begin{abstract}
Our paper addresses the problem of models struggling to learn diverse features, due to either forgetting previously learned features or failing to learn new ones. To overcome this problem, we introduce Diverse Feature Learning (DFL), a method that combines an important feature preservation algorithm with a new feature learning algorithm. Specifically, for preserving important features, we utilize self-distillation in ensemble models by selecting the meaningful model weights observed during training. For learning new features, we employ reset that involves periodically re-initializing part of the model. As a result, through experiments with various models on the image classification, we have identified the potential for synergistic effects between self-distillation and reset.
\end{abstract}

\section{Introduction}
\label{sec:intro}
To solve a task, it is important to know the related features. For example, in colorization, proper segmentation features are necessary to color in the correct locations \cite{hicsonmez2023improving, yang2022bistnet}. However, in deep learning, there is an issue with feature learning related to the phenomenon of forgetting learned features \cite{zhou2022fortuitous} or not being able to learn new features \cite{park2023training}. In other words, there is a need for a method that can preserve learned features while learning new features. Therefore, we propose Diverse Feature Learning (DFL), a method that can take the advantages of both feature preservation algorithms and new feature learning algorithms.

For the feature preservation algorithm, we implement self-distillation on ensemble models based on the training trajectory. This approach assumes that the model retains knowledge about important features throughout training but can also forget them. Therefore, by properly selecting models on the training trajectory and applying self-distillation, we leverage the advantage of the alignment of important features, ensuring their preservation.

For the new feature learning algorithm, we have utilized reset, which involves periodically re-initializing part of the model. This strategy is based on the assumption that learning with gradient descent is confined to a limited weight space, potentially limiting the learning of specific features. Under this assumption, reset allows the model to explore different constrained weight spaces, thereby enabling the learning of new features.

Our DFL is completed by combining self-distillation and reset. Through this combination, it creates a synergistic effect that enables the learning of diverse important features. The effectiveness of our algorithm has been experimentally demonstrated using the image classification on CIFAR-10 and CIFAR-100 \cite{krizhevsky2009learning}.

Our contributions are as follows.

\begin{enumerate}
  \item We propose Diverse Feature Learning (DFL), a novel approach that synergistically combines feature preservation through self-distillation with new feature learning via reset.
  \item The proposed algorithm has been experimentally shown potential to aid model performance on the image classification.
\end{enumerate}

The structure of the paper is as follows: To facilitate understanding of the reasons behind our method and explain the details of the algorithm, we discuss the related work in Section~\ref{sec:rel} and our method in Section~\ref{sec:method}. Subsequently, we demonstrate how our model is beneficial through experiments in Section~\ref{sec:exp}. Finally, we conclude with our limitations and future work in Section~\ref{sec:con}.

\section{Related Work}
\label{sec:rel}
In this session, we explain the importance of learning various features and the insight behind the development of our algorithm in Section~\ref{sec:div}. Then, considering that our Diverse Feature Learning (DFL) is primarily composed of two key components: feature preservation and new feature learning, we outline the differences between our algorithm and existing approaches in Sections~\ref{sec:pre} and~\ref{sec:new}.

\subsection{Diverse Feature}
\label{sec:div}
The ability to learn various features contributes to the improved performance of deep learning models \cite{nicolicioiu2023learning}. A prime example of how utilizing diverse features can be beneficial is the enhancement of accuracy and robustness in ensembles, which are based on using models with distinct hyperparameters \cite{ganaie2022ensemble}. Because it has been reported that ensemble methods are more effective when the errors between different models are uncorrelated \cite{Tseng2020Cross-Domain}.

While it has been proven that learning various features in an ensemble yields positive effects, there are challenges associated with learning various features in the context of single model training. For example, learning various features can be hindered by interpreting the same feature differently \cite{choi2019self}, and by perceiving different features in the same way because of spurious correlations \cite{perera2023analyzing, schramowski2023safe, choi2022perception} and as well as issues resulting from multi-view \cite{allen-zhu2023towards}. 

Efforts have been made to solve the obstacles in learning features by focusing on data-centric approaches. For example, resampling to adjust the distribution of training data has been utilized to concentrate on learning key features \cite{li2020background, teterwak2023erm++}. Additionally, identifying main features has been facilitated by training different models on various training sets \cite{wen2021debiased, kim2022learning}. However, methods that exclude certain data or involve separate training can have the drawback of losing information compared to approaches that train on all data at once.

In summary, it has been observed that learning diverse features in ensembles is beneficial. However, training with a single model can lead to the learning of unwanted features, which hinders the learning of diverse features. In relation to this, there are efforts focused on resolving this issue primarily through data-centric approaches. Nevertheless, data processing can result in the loss of information. Therefore, we aimed to incorporate the characteristics of ensembles into the training process of a single model, based on the themes of feature preservation and new feature learning, to enable the learning of diverse features even in single model training.

\subsection{Feature Preservation}
\label{sec:pre}
In deep learning, feature preservation involves the issue of forgetting existing features while learning new ones \cite{ke2022continual}. To solve this, pre-trained weights could be stochastically restored \cite{Wang_2022_CVPR}, or training could focus on only certain layers \cite{ke2022continual, kirichenko2023last, park2023training}. Additionally, the training process includes aligning features \cite{jung2023cafa}. This can be considered an attempt to utilize previously learned features. However, simply utilizing existing features can be disadvantageous, as it might also preserve features that are not meaningful.

To address this issue, we sought to develop an algorithm based on the assumption that important features, shared among features learned differently, can be preserved by utilizing multiple sets of weights, drawing inspiration from ensembles. In other words, we aimed to use significant weights on the training trajectory as the teacher model for self-distillation. This approach has similarities to previous work that created a teacher model based on exponential moving average of the training trajectory weights \cite{tarvainen2017mean}, the current weights \cite{zhang2019your},  or the weights of the immediate previous epoch \cite{kim2021self}. However, our method is distinct: we select weights to be used for self-distillation from the learning process, which is motivated by the idea that ensembles using only significant weights based on validation performance can lead to additional performance improvement \cite{wortsman2022model}.

\subsection{New Feature Learning}
\label{sec:new}
Learning new features can be beneficial for improving generalization \cite{montero2020role}. However, difficulties in feature learning may arise depending on the distribution of the model's weights \cite{nikishin2023deep}. Considering the relationship between weights and learning difficulties, we can hypothesize that once a deep learning model learns certain features, it might become biased towards learning other features in a specific, potentially non-generalized direction. This tendency could occur because deep learning models have enough degrees of freedom to learn features, even from noise \cite{zhang2021understanding}. In other words, this hypothesis suggests that certain useful features are difficult to learn through gradient descent when started from specific weights.

Based on this hypothesis, methods like random projection and model re-initialization have effectively addressed challenges in learning useful features. Random projection mitigates the issue of biased learning towards certain misaligned features \cite{sauer2021projected}. In the domain of reinforcement learning, reset, which re-initializes the model, has proven to be beneficial for training without losing plasticity \cite{nikishin2023deep}. Among these methods, our algorithm incorporates reset to facilitate the learning of new features, as the adjustment of weights could be effectively integrated with the feature preservation strategy that enforces a consistency loss on weights.

In summary, we explain the rationale behind our design of DFL, which integrates self-distillation and reset. The motivation can be summarized as bringing the advantages of an ensemble to the training of a single model. Specifically, we select the teachers for self-distillation from the training trajectory to leverage the advantages of utilizing various weights. Furthermore, drawing inspiration from the benefit of uncorrelated errors in ensemble models, we employed reset to learn new features.

\section{Methods}
\label{sec:method}
In this section, we describe the general algorithm of our proposed Diverse Feature Learning (DFL), as well as the algorithm used in our experiment. The proposed learning algorithm is detailed in Section~\ref{sec:alg_gen}. Then, we discuss the algorithm specifically tailored for the experiment in Section~\ref{sec:alg_cls}.

\subsection{General Algorithm}
\label{sec:alg_gen}
Our DFL is based on self-distillation and reset as described in Algorthm~\ref{alg:general}. It involves selecting certain layers of the model as the student and applying self-distillation to them. The teacher, which are targeted for self-distillation, are the same architecture as the student and are updated with past weights from the training trajectory of the student. In other words, multiple teachers have been appointed, and the train process includes replacing the least meaningful teacher with the student, based on a meaningfulness measurement. Additionally, to facilitate learning new features, we do reset the student which means periodically re-initialize the student.

To elaborate, in deep learning, we can assume that we have a dataset \( S \) with \( N \) samples, comprising inputs \( X = \{x^i\}_{i=1}^N \) and outputs \( Y = \{y^i\}_{i=1}^N \). With this dataset, we train the model \( \theta \). A part of this model \( \theta \) is selected as the student \( \phi_0 \). Then, the teacher set \( \Phi = \{\phi_k\}_{k=1}^K \) comprises \( K \) multiple teachers. Additionally, each teacher \( \phi_k \) consists of layers identical to those of the student \( \phi_0 \). Since all instances of \( \phi_k \) share the same architecture, it is possible to apply different \( \phi_k \) to the model \( \theta \). Depending on which \( \phi_k \) is utilized, \( \theta \) is accordingly represented as \( \theta_k \).

At each step \( t \), with the sampled data \((x^t, y^t)\), we predict \(\hat{y}^t_k\) using each \(\theta_k\). During this prediction process, we also obtain the output \(\hat{z}^t_k\) from either the student $\phi_0$ or each teacher $\phi_k$. Subsequently, meaningfulness \(p_k\) of each model $\theta_k$ is calculated based on the meaningfulness measurement function \(f\). This function \(f\) is dependent on the previous meaningfulness \(p_k\), the prediction \(\hat{y}_k^t\), and the actual output \(y^t\). Based on this calculated meaningfulness \(p_k\), the teacher set \(\Phi\) is updated in accordance with the teacher update cycle \( T_{\text{update}} \). Specifically, if the student's meaningfulness \( p_0 \) is equal to or higher than the least meaningfulness \( p_{k'= \arg \min_{k \in [1, K]} p_k}  \) in the teacher set $\Phi$, the least meaningful teacher \( \phi_{k'} \) is replaced with the student \( \phi_0 \).

Also, at each step \( t \), the model with the student \( \theta_0 \) is trained using two types of losses: the task-specific loss \( \mathcal{L}_{\text{main}} \) and the self-distillation loss \( \mathcal{L}_{\text{distill}} \). The self-distillation loss \( \mathcal{L}_{\text{distill}} \) is calculated based on the consistency loss between the student's feature \( \hat{z}^t_0 \) and each teacher's feature \( \hat{z}^t_k \). Additionally, to encourage diverse learning of the student $\phi_0$, an update occurs at every student reset cycle \( T_{\text{reset}} \), which involves re-initializing the student $\phi_0$.

\begin{algorithm}[ht]
   \caption{Diverse Feature Learning (DFL)}
   \label{alg:general}
\begin{algorithmic}
   \STATE {\bfseries Input:} dataset $S \stackrel{\triangle}{=}\{(x^i, y^i)\}^N_{i=1}$
   \STATE {\bfseries Input:} model $\theta$ includes student $\phi_0 \subset \theta$
   \STATE {\bfseries Input:} teacher set $\Phi = \{\phi_k\}_{k=1}^K$
   \STATE {\bfseries Input:} meaningfulness measurement $f$
   \STATE {\bfseries Input:} student and teachers' meaningfulness $P = \{p_k\}_{k=0}^K$
   \STATE {\bfseries Input:} teacher update cycle $T_{\text{update}}$
   \STATE {\bfseries Input:} student reset cycle $T_{\text{reset}}$
   \STATE {\bfseries Input:} loss function $\mathcal{L}_{\text{main}}, \mathcal{L}_{\text{distill}}$
   \STATE {\bfseries Input:} learning rate $\alpha$, current step $t$
   \STATE Initialize $t, P, \theta, \Phi$
   \REPEAT
        \STATE $t \leftarrow t + 1$
        \STATE $(x^t,y^t) \sim S$
        \FOR{$k=0$ {\bfseries to} $K$}
            \STATE $\theta_k \leftarrow \theta$ with $\phi_k$
            \STATE $\hat{y}_k^t, \hat{z}_k^t \leftarrow \theta_k(x^t)$
            \STATE $p_k \leftarrow f(p_k, \hat{y}_k^t, y^t)$
        \ENDFOR
        \STATE $\mathcal{L}_{\text{total}} = \mathcal{L}_{\text{main}}(\hat{y}^t_0, y^t) + \sum^K_{k=1} \mathcal{L}_{\text{distill}}(\hat{z}_0^t, \hat{z}_k^t)$
        \STATE $\theta_0 \leftarrow \theta_0 - \alpha \nabla_{\theta_0} \mathcal{L}_{\text{total}}$
        
        \IF{$t \bmod T_{\text{update}} = 0$}
            \STATE $k' \leftarrow \arg \min_{k \in [1, K]} p_k$
            \IF{$p_{k'} \le p_0$}
                \STATE Update: $\phi_{k'} \leftarrow \phi_0$
            \ELSE
                \STATE Update: $\phi_{k'} \leftarrow \phi_{k'}$
            \ENDIF
            \STATE Re-initialize $P$
        \ENDIF
        \IF{$t \bmod T_{\text{reset}} = 0$}
            \STATE Re-initialize $\phi_0$
        \ENDIF
   \UNTIL{$\theta_0$ not converged}
\end{algorithmic}
\end{algorithm}

To optimize practical use, decisions on key aspects are required. For self-distillation, these include choosing the layers of the model for the student $\phi_0$, determining the number of teachers $K$, selecting a function $f$ to measure meaningfulness, and setting the duration of the teacher update cycle $T_{\text{update}}$. For reset, it is necessary to decide on the duration of the student reset cycle $T_{\text{reset}}$ and the method of re-initialization.

\subsection{Algorithm for Image Classification}
\label{sec:alg_cls}
For the image classification, we have a dataset \(S\) with \(N\) samples, comprising images \(X = \{x^i\}_{i=1}^N\) and corresponding labels \(Y = \{y^i\}_{i=1}^N, y^i \in \{1, 2, ..., M\}\) with \(M\) classes.

\begin{figure}[t]
\vskip 0.2in
\begin{center}
\centerline{\includegraphics[width=0.9\columnwidth]{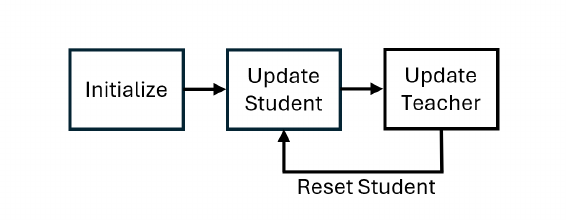}}
\caption{\textbf{Overall Training Process} Our algorithm includes the processes of updating the student through self-distillation for feature preservation, updating the teacher based on meaningfulness for efficient ensemble, and updating the student with reset for new feature learning.}
\label{fig:total}
\end{center}
\vskip -0.2in
\end{figure}

Then, the total process for training is as shown in Figure~\ref{fig:total}. This process aligns with Algorithm~\ref{alg:general}, utilizing the identical cycle \(T_{\text{cycle}}\) for both the teacher update cycle \(T_{\text{update}}\) and the student reset cycle \(T_{\text{reset}}\). In detail, after initializing the model, the process involves repeatedly updating the student, followed by the teacher. Subsequently, we will sequentially explain each process used in our experiments.

\begin{figure}[t!]
\vskip 0.2in
\begin{center}
\centerline{\includegraphics[width=0.9\columnwidth]{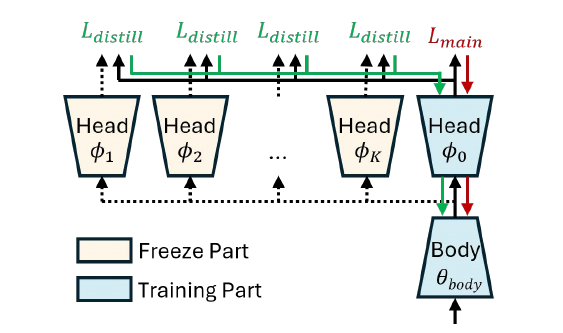}}
\caption{\textbf{Update Student} Training the model $\theta$ involves using the classification loss \( \mathcal{L}_{\text{main}} \) and self-distillation loss \( \mathcal{L}_{\text{distill}} \). With these losses, the student \(\phi_0\) and the body \(\theta_{\text{body}}\) are updated, while the teacher set \(\Phi = \{\phi_k\}_{k=1}^K\) is kept frozen.} 
\label{fig:student}
\end{center}
\vskip -0.2in
\end{figure}

\textbf{Initialize} We utilize the last layer or the last part of layers as the student \(\phi_0\) in the model \(\theta\), which facilitates training with minimal computational difference. Consequently, each teacher \(\phi_k \in \Phi = \{\phi_k\}_{k=1}^K\) mirrors the student's structure, as depicted in Figure~\ref{fig:student}. As a result, we initialize both the model \(\theta\) and the teacher set \(\Phi\). Since we use the last part of the layers, this part is referred to as the head. Furthermore, in model \(\theta\), the parts other than the head are referred to as the body \(\theta_{\text{body}}\).

\textbf{Update Student} Next, we will describe the process of updating the student, as shown in Figure~\ref{fig:student}. It utilizes the classification loss \(\mathcal{L}_{\text{main}}\) and the self-distillation loss \(\mathcal{L}_{\text{distill}}\). Based on these losses, the student $\phi_0$ and the body $\theta_{\text{body}}$ are trained, but the teacher set $\Phi$ is frozen, meaning its weights are not updated.

The updating of the student proceeds as follows: The image \(x\) is first processed through the body \(\theta_{\text{body}}\). Then, it passes through the student and all the teachers \(\{\phi_k\}_{k=0}^K\). Subsequently, for each output, a prediction \(\hat{y}\) for the label \(y\) is obtained using softmax. At the training step \(t\), this process can be mathematically represented as shown in Equation~\ref{eq:pred}.

\begin{equation}
\begin{aligned}
\label{eq:pred}
    q_k^t &= \phi_k(\theta_{\text{body}}(x^t)), \\
    \hat{y}_k^t &= \text{Softmax}(q_k^t).
\end{aligned}
\end{equation}

Then, the classification loss \(\mathcal{L}_{\text{main}}\) is defined with the cross-entropy as shown in Equation~\ref{eq:main}. Additionally, self-distillation loss \(\mathcal{L}_{\text{distill}}\) utilizes the Kullback–Leibler (KL) divergence between the student $\phi_0$ and each teacher $\phi_k$, as detailed in Equation~\ref{eq:class}. Consequently, the total loss is formulated in Equation~\ref{eq:total}.

\begin{align}
    \mathcal{L}_{\text{main}}(\hat{y}_0^t, y^t) &= \text{CrossEntropy}(\hat{y}_0^t, y^t) \label{eq:main}\\
    \mathcal{L}_{\text{distill}}(\hat{y}_0^t, \hat{y}_k^t) &= \text{KL} (\hat{y}_0^t, \hat{y}_k^t) \label{eq:class} \\
    \mathcal{L}_{\text{total}} = \mathcal{L}_{\text{main}}(\hat{y}_0^t, y^t) &+ \frac{1}{K} \sum_{k=1}^K \mathcal{L}_{\text{distill}}(\hat{y}_0^t, \hat{y}_k^t) \label{eq:total}
\end{align}

\begin{figure}[ht!]
\vskip 0.2in
\begin{center}
\centerline{\includegraphics[width=0.9\columnwidth]{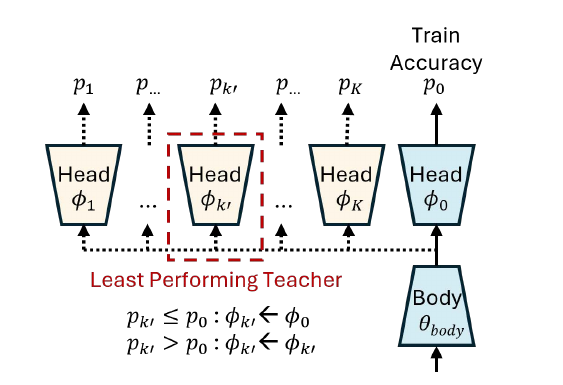}}
\caption{\textbf{Update Teacher} Updating the teacher set \(\Phi\) involves replacing the least performing teacher \(\phi_{k'}\) with the student \(\phi_0\) if the former underperforms. This determination is based on the meaningfulness measurement $f$. For the image classification, we measure the meaningfulness \(p_k\) with the training accuracy in the most recent epoch.}
\label{fig:teacher}
\end{center}
\vskip -0.2in
\end{figure}

\textbf{Update Teacher} To update the teacher, we measure the meaningfulness \(p_k\) of all heads and retain the better ones, as shown in Figure~\ref{fig:teacher}. For the image classification, the training accuracy from the most recent epoch serves as our measure of meaningfulness \(p_k\). When it's time to update the teacher based on meaningfulness \(p_k\), the least performing teacher \(\phi_{k'}\), from the teacher set \(\Phi\), is compared with the student \(\phi_0\). If the student \(\phi_0\) performs the same as or better than the teacher \(\phi_{k'}\), the weights of the teacher \(\phi_{k'}\) are updated with those of the student \(\phi_0\). Conversely, if the student \(\phi_0\) performs worse, no update is carried out. In other words, from the combined set of the teacher set \(\Phi\) and the student \(\phi_0\), the top-\(K\) performing heads are selected to form the new teacher set \(\Phi\). This process can be described using Equation~\ref{eq:update}.

\begin{equation}
\begin{aligned}
\label{eq:update}
    \hat{Y}_k &= \{\hat{y}_k^i\}_{i=1}^N \\
    p_k &= \text{Accuracy}(\hat{Y}_k, Y) \\
    k' &= \arg \min_{k \in [0, K]} p_k \\
    \phi_{k'} &\leftarrow \phi_0
\end{aligned}
\end{equation}

\textbf{Reset Student} Finally, we explain the process of updating the student $\phi_0$ through reset. We utilized two reset methods: random reset and mean reset. Random reset involves re-initializing the student $\phi_0$ with random weights, while mean reset involves re-initializing the student \(\phi_0\) based on the mean weights of the teacher set \(\Phi\), as shown in Equation~\ref{eq:reset}. We experimented with mean reset because we hypothesized that it helps to prevent the student $\phi_0$ from learning the same features in different ways, as compared to the teacher set \(\Phi\).

\begin{align}
\label{eq:reset}
    \phi_0 &\leftarrow \frac{1}{K} \sum_{k=1}^K \phi_k
\end{align}

In summary, our DFL approach for the experiment is described in Algorithm~\ref{alg:cls}. We use the last part of the model, referred to as the head, for the student $\phi_0$ and the teacher set $\Phi$ in our algorithm. Additionally, we use the training accuracy of the last epoch as the measure of meaningfulness. Furthermore, for reset, we employ random reset and mean reset. Then, the identical cycle \(T_{\text{cycle}}\) is used for both the teacher update cycle \(T_{\text{update}}\) and the student reset cycle \(T_{\text{reset}}\).

\begin{algorithm}[ht]
   \caption{DFL for Image Classification}
   \label{alg:cls}
\begin{algorithmic}
   \STATE {\bfseries Input:} dataset $S \stackrel{\triangle}{=}\{(x^i, y^i)\}^N_{i=1}$
   \STATE {\bfseries Input:} body $\theta_{\text{body}}$, student $\phi_0$
   \STATE {\bfseries Input:} teacher set $\Phi = \{\phi_k\}_{k=1}^K$
   \STATE {\bfseries Input:} previous epoch accuracy measurement $f$
   \STATE {\bfseries Input:} student and teachers' epoch accuracy $\{p_k\}_{k=0}^K$
   \STATE {\bfseries Input:} teacher update and student reset cycle $T_{\text{cycle}}$
   \STATE {\bfseries Input:} loss function $\mathcal{L}_{\text{main}}, \mathcal{L}_{\text{distill}}$
   \STATE {\bfseries Input:} learning rate $\alpha$, current step $t$
   \STATE Initialize $t, \theta_{\text{body}}, \phi_0, \Phi$
   \REPEAT
        \STATE $t \leftarrow t + 1$
        \STATE $(x^t,y^t) \sim S$
        \FOR{$k=0$ {\bfseries to} $K$}
            \STATE $\hat{y}_k^t \leftarrow \phi_k(\theta_{\text{body}}(x^t))$
            \STATE $p_k \leftarrow f(t, p_k, \hat{y}_k^t, y^t)$
        \ENDFOR
        \STATE $\mathcal{L}_{\text{total}} = \mathcal{L}_{\text{main}}(\hat{y}^t_0, y^t) + \sum^K_{k=1} \mathcal{L}_{\text{distill}}(\hat{y}_0^t, \hat{y}_k^t)$
        \STATE $(\theta_{\text{body}}, \phi_0) \leftarrow (\theta_{\text{body}}, \phi_0) - \alpha \nabla_{(\theta_{\text{body}}, \phi_0)} \mathcal{L}_{\text{total}}$
        \IF{$t \bmod T_{\text{cycle}} = 0$}
            \STATE $k' \leftarrow \arg \min_{k \in [1, K]} p_k$
            \IF{$p_{k'} \le p_0$}
                \STATE Update: $\phi_{k'} \leftarrow \phi_0$
            \ELSE
                \STATE Update: $\phi_{k'} \leftarrow \phi_{k'}$
            \ENDIF
            \STATE Re-initialize $\phi_0$
        \ENDIF
   \UNTIL{$(\theta_{\text{body}}, \phi_0)$ not converged}
\end{algorithmic}
\end{algorithm}

\section{Experiment}
\label{sec:exp}
To demonstrate the effectiveness of our Diverse Feature Learning (DFL), we experiment with various model architectures in the image classification. Then, we explain how the components of DFL, feature preservation and new feature learning, have the potential to improve performance in the image classification.

\subsection{Dataset}
We utilized the CIFAR-10 and CIFAR-100 datasets for the image classification \cite{krizhevsky2009learning}. The CIFAR-10 dataset contains 60,000 images, each with a resolution of 32x32 pixels, and comprises 10 labels: airplane, automobile (excluding trucks), bird, cat, deer, dog, frog, horse, ship, and truck. There are 5,000 training images and 1,000 test images for each label. The dataset is filtered to ensure the inclusion of only photo-realistic images, excluding any that resemble line drawings. Data featuring occlusions or unusual viewpoints are included if they are identifiable. In all aspects except for the number of labels and the associated images per label, the CIFAR-100 dataset is identical to CIFAR-10. CIFAR-100 contains 100 labels, each with 500 training images and 100 test images.

\subsection{Model Architecture}
We conducted experiments on five lightweight models using a GeForce 2080Ti and a TITAN RTX GPU. The models include a VGG-16 \cite{simonyan2015very}, a Squeezenet \cite{iandola2016squeezenet}, a Shufflenet \cite{zhang2018shufflenet}, a Mobilenet-Version-2 \cite{sandler2018mobilenetv2}, and a GoogLeNet  model \cite{szegedy2015going}. Each model, based on a convolutional neural network, has slight differences in its specific architecture.

To briefly describe the architecture of each model: The VGG model features a convolutional neural network with 3x3 filters. The SqueezeNet model employs a Fire module, combining 1x1 and 3x3 convolution. The ShuffleNet model uses channel shuffling and depthwise separable convolution. The MobileNet model utilizes an inverted residual structure with linear bottlenecks, 1x1 convolution, and depthwise separable convolution. Lastly, the GoogLeNet  model incorporates an Inception module where convolution filters of different sizes and max pooling are placed in parallel.

\subsection{Hyperparameter}
We use Stochastic Gradient Descent (SGD) with momentum as the optimizer. The momentum \(\mu\) was set to 0.9, and the learning rate \(lr\) to 0.1. A warm-up step was implemented for 1 epoch as the learning rate scheduler, followed by multiplying the learning rate by 0.2 at epochs 60, 120, and 160. The training consisted of a total of 200 epochs, and each step was conducted with a batch size of 128. Each of these hyperparameters utilizes the default settings of the baseline code \cite{weiaicunzai_pytorch_cifar100} and the paper \cite{devries2017improved}. The weight decay \(\lambda\) for SGD was set to 2e-4.

\subsection{Results}
In this session, we present the results related to DFL as outlined in Algorithm~\ref{alg:cls}. First, we investigate the significance of DFL for the VGG model on CIFAR-100. Then, we explore the characteristics of DFL by varying hyperparameters and model architectures. Each analysis is based on a grid search for the number of teachers $K$ at $\{1, 2, 4, 8\}$ and the cycle $T_{\text{cycle}}$ at $\{1, 20, 50, 100\}$. Overall experimental results can be seen in Appendix~\ref{app}.

\textbf{Significance of DFL} To demonstrate the significance of DFL, we present an analysis of the results on CIFAR-100 using the VGG model. The configuration parameters for DFL were adopted from the best settings for CIFAR-10. For DFL, the student was defined as the last three layers. The number of teachers \(K\) for self-distillation was set to 4. Additionally, the cycle \(T_{\text{cycle}}\) was set to 100 epochs. Furthermore, reset was carried out based on the average of the teachers for re-initialization.

In Table~\ref{tbl:algorithm}, the results based on the above settings are as follows: When only reset was applied, the mean accuracy decreased by 0.14\%. When training with self-distillation based on the weights of the previous epoch, the accuracy increased by 0.56\%. Training with self-distillation based on random initialization, the accuracy increased by 0.36\%. And in the case of DFL utilizing both reset and self-distillation, there was a 1.09\% increase.

The 0.14\% decrease in accuracy indicates that resetting the last part of layers in CIFAR-100 has not positive impact on performance, thus reproducing the findings \cite{zhou2022fortuitous} in our experimental setting. Additionally, our results also reproduce the findings \cite{kim2021self}, demonstrating that self-distillation based on the previous epoch contributes to a slight improvement in accuracy of 0.56\%. Furthermore, the increase in accuracy of 0.36\% achieved by incorporating self-distillation with random initialization suggests that the performance enhancements observed in reinforcement learning through distillation of random networks \cite{burda2018exploration}, can also be applicable to image classification.

Our DFL shows the highest performance improvement of 1.09\%. This is because we combines reset and self-distillation appropriately based on feature preservation and new feature learning. In other words, while reset has the disadvantage of being unrelated to existing features as it explores weights for new feature learning, this issue is resolved by applying self-distillation. Moreover, in terms of feature preservation, comparing with two other types of self-distillation, it is also beneficial that the teacher can be selected in the training trajectory based on measuring meaningfulness.

\begin{table}[H]
\caption{\textbf{Self-distillation and Reset} It represents the mean accuracy and standard deviation for the combination of reset and self-distillation with the VGG model on CIFAR-100. The mean and standard deviation were calculated based on five different seeds. The self-distill in the table refers to self-distillation.}
\label{tbl:algorithm}
\vskip 0.15in
\begin{center}
\begin{small}
\begin{sc}
\begin{tabular}{lr}
\toprule
Algorithm & Accuracy \\
\midrule
default                                  & 71.85 ± 0.32 \\
\; + reset                           & 71.71 ± 0.28 \\
\; + self-distill (prev epoch)       & 72.41 ± 0.23 \\
\; + self-distill (rand)             & 72.21 ± 0.25 \\
\; + DFL (reset + self-distill) & \textbf{72.94 ± 0.44} \\
\bottomrule
\end{tabular}
\end{sc}
\end{small}
\end{center}
\vskip -0.1in
\end{table}

\textbf{Characteristics of DFL} Based on CIFAR-100 using the VGG model, we analyze the characteristics of DFL by varying the target layer of the student, the duration of the cycle $T_{\text{cycle}}$, the number of teachers $K$, and the method of reset. For the target layer of the student, we compared the results based on how many layers were utilized, particularly focusing on the head part, which is the last part of the model. Additionally, for the analysis of the duration of the cycle $T_{\text{cycle}}$ and the number of teachers, we examined the results from the grid search. Finally, for reset, we compared methods of random re-initialization and re-initialization based on the mean of the teachers.

According to Table~\ref{tbl:ablation} with number of layer $L$ part, we can confirm that defining the student with three last layers, as opposed to just one, results in an accuracy improvement of 0.56\%. This suggests that increasing the number of layers that define the student can offer sufficient flexibility for reset to learn new features, which could be beneficial for DFL.

The following experiment concerns the duration of the cycle $T_{\text{cycle}}$ in terms of epochs. The results are presented in Table~\ref{tbl:ablation} for the duration of cycle $T$. Using DFL with a cycle of 50 epochs results in a performance improvement of 0.71\%, while using DFL with a cycle of 100 epochs leads to an improvement of 1.09\%. For shorter cycles, there was a 0.25\% performance improvement for 1 epoch and a 0.22\% performance decrease for 20 epochs.

This means that DFL can show performance improvement when there is an appropriate duration for the cycle \(T_{\text{cycle}}\). There are two possible reasons why an appropriate duration for the cycle \(T_{\text{cycle}}\) is necessary. The first is that sufficient iteration is needed after reset. This is because it is essential to update the student that could utilize the features of the body.

Another reason relates to using the previous epoch's training accuracy as a meaningfulness measurement. While this approach effectively selects a more significant teacher with better-aligned features than the initial weights, issues could arise when assessing the trained model's meaningfulness. This is because if the model is prone to overfitting, the student might not be adequately trained to utilize the aligned features, yet could still exhibit high accuracy. Consequently, such a student would become a teacher. This suggests that one of DFL's key goals, namely feature preservation through commonalities among models, is not being met. Therefore, in cases with a cycle duration exceeding 50 epochs, it can be said that performance has stably improved due to fewer updates. In other words, fewer updates are mostly advantageous as they tend to better update the teacher set compared to the initial teacher set.

Regarding the number of teachers \( K \) and the re-initialization algorithm for reset \( M \) in Table~\ref{tbl:ablation}, the changes in performance are as follows. The performance improvements depending on the number of teachers showed increases of 1.04\%, 1.18\%, and 1.09\% for one, two, and four teachers, respectively. However, when increasing to eight teachers, there is a performance decline of 3.85\%. Regarding the differences in the re-initialization algorithm, using random reset or mean reset resulted in performance improvements of 1.19\% and 1.09\%, respectively.

The analysis of these results suggests that DFL demonstrates a consistently meaningful performance improvement when varying hyperparameters and applying different re-initialization algorithms for reset. However, the minor differences in performance improvement among each hyperparameter could be attributed to the limitations of the meaningfulness measurement, as analyzed in the duration of cycle \( T_{\text{cycle}} \). In other words, by using the previous epoch's training accuracy as the meaningfulness measurement, updating the teacher set compared to the initial teacher set ensures that the teachers have more aligned features. However, this does not guarantee continued improvement beyond that point.

In this context, as the number of teachers increases, the potential advantage of observing more existing features is not realized. In fact, a performance decline is observed when the number of teachers is large, such as eight. This could be due to the short total training duration, resulting in insufficient updates of the teacher set from its initial state. Similarly, the minimal performance differences observed between various re-initialization methods for reset could be attributed to the low alignment of features among teachers.

\begin{table}[H]
\caption{\textbf{Ablation Study on CIFAR-10 and CIFAR-100} It represents the mean accuracy and standard deviation with the VGG model. The mean and standard deviation were calculated based on five different seeds. L represents the number of layers utilized by the student. K signifies the number of teachers utilized. T stands for the the duration of cycle \( T_{\text{cycle}} \). M indicates the re-initialization method used for reset. When M is marked as O, it means that re-initialization was done using the mean of the teachers' weights. X signifies that random weights were utilized for re-initialization. The notation - signifies the use of the default setting. The default setting for (L, K, T, M) is (3, 4, 100, O).}
\label{tbl:ablation}
\vskip 0.15in
\begin{center}
\begin{small}
\begin{sc}
\begin{tabular}{lcccccr}
\toprule
L & K & T & M & CIFAR-10 & CIFAR-100 \\
\midrule
X & X & X & X & 93.64 ± 0.14 & 71.85 ± 0.32 \\
3 & 4 & 100 & O & \textbf{93.78 ± 0.10} & \textbf{72.94 ± 0.44} \\
\hline
1 & - & - & - & 93.49 ± 0.31 & 72.38 ± 0.18 \\
3 & - & - & - & \textbf{93.78 ± 0.78} & \textbf{72.94 ± 0.44} \\
\hline
- & 1 & - & - & 93.71 ± 0.12 & 72.89 ± 0.33 \\
- & 2 & - & - & 93.77 ± 0.18 & \textbf{73.03 ± 0.24} \\
- & 4 & - & - & \textbf{93.78 ± 0.10} & 72.94 ± 0.24 \\
- & 8 & - & - & 93.22 ± 0.13 & 68.00 ± 2.09 \\
\hline
- & - & 1   & - & 93.53 ± 0.12 & 72.10 ± 0.30 \\
- & - & 20  & - & 93.48 ± 0.15 & 71.63 ± 0.34 \\
- & - & 50  & - & 93.61 ± 0.06 & 72.56 ± 0.22 \\
- & - & 100 & - & \textbf{93.78 ± 0.10} & \textbf{72.94 ± 0.44} \\
\hline
- & - & - & X & 93.64 ± 0.14 & \textbf{73.04 ± 0.09} \\
- & - & - & O & \textbf{93.78 ± 0.10} & 72.94 ± 0.44 \\
\bottomrule
\end{tabular}
\end{sc}
\end{small}
\end{center}
\vskip -0.1in
\end{table}

\begin{figure}[t]
\vskip 0.2in
\begin{center}
\centerline{\includegraphics[width=0.86\columnwidth]{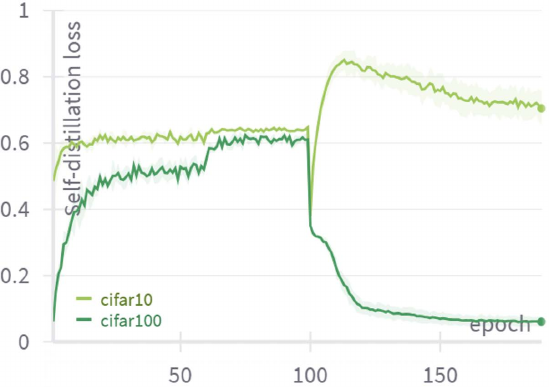}}
\caption{\textbf{Self-distillation Loss over Different Dataset} It represents the mean self-distillation loss and standard deviation using the default settings of Table~\ref{tbl:ablation}. The lines of the graph represent the mean, while the shaded areas indicate the standard deviation.}
\label{fig:dataset}
\end{center}
\vskip -0.2in
\end{figure}

Table~\ref{tbl:ablation} presents similar characteristics between CIFAR-10 and CIFAR-100. However, CIFAR-10 exhibits a smaller performance difference compared to CIFAR-100. This difference can be attributed to CIFAR-10's fewer classes, making it more vulnerable to overfitting. This vulnerability is depicted by the training loss curve in Figure~\ref{fig:default_dataset} located in Appendix~\ref{app}. Consequently, this leads to a reduced correlation between updated teachers, making it difficult to decrease the self-distillation loss $\mathcal{L}_{\text{distill}}$, as illustrated in Figure~\ref{fig:dataset}.

\textbf{Diverse Model Architecture} We analyze whether our DFL benefits the improvement of results across five different models, as presented in Table~\ref{tbl:diverse}. It shows the results when the student model is defined with only the last layer. The VGG model exhibited a performance improvement of 0.53\%, and the Squeezenet model showed a 0.61\% improvement. The Shufflenet model demonstrated a 0.73\% improvement. However, there was a slight decline of 0.16\% for the Mobilenet model, and a decrease of 0.58\% for the GoogLeNet model.

\begin{figure}[t]
\vskip 0.2in
\begin{center}
\centerline{\includegraphics[width=0.9\columnwidth]{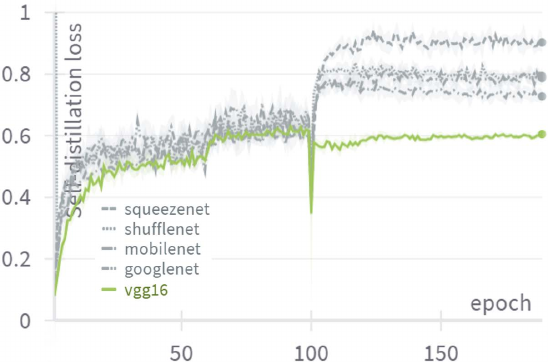}}
\caption{\textbf{Self-distillation Loss over Different Model Architecture} It represents the mean self-distillation loss and standard deviation using the setting of Table~\ref{tbl:diverse}. The lines of the graph represent the mean, while the shaded areas indicate the standard deviation.}
\label{fig:architecture}
\end{center}
\vskip -0.2in
\end{figure}

The results indicate that the difference is relatively small compared to the data in Table~\ref{tbl:algorithm}. This could be due to the fact that the self-distillation loss, as depicted in Figure~\ref{fig:architecture}, does not decrease significantly, especially when compared to the dark green line in Figure~\ref{fig:dataset}, which employs more layers for the student. This suggests that students with only the last layer of the model face challenges in learning aligned features through both the classification loss and the self-distillation loss from various teachers. In other words, when the student's capacity is limited, it becomes challenging for them to learn effectively.

\begin{table}[H]
\caption{\textbf{Diverse Model Architecture on CIFAR-100} It represents the mean accuracy and standard deviation for each of the five models, tested with five different seeds. The last layer of the model is set as the student, and other parameters follow the default settings outlined in Table~\ref{tbl:ablation}. Text in bold indicates the highest mean accuracy.}
\label{tbl:diverse}
\vskip 0.15in
\begin{center}
\begin{small}
\begin{sc}
\begin{tabular}{lcr}
\toprule
Model(L=1) & Without DFL & With DFL \\
\midrule
vgg         & 71.85 ± 0.32 & \textbf{72.38 ± 0.18} \\
squeezenet  & 69.53 ± 0.41 & \textbf{70.14 ± 0.22} \\
shufflenet  & 70.45 ± 0.35 & \textbf{71.18 ± 0.59} \\
mobilenet   & \textbf{70.30 ± 0.29} & 70.14 ± 0.43 \\
googlenet   & \textbf{76.52 ± 0.39} & 75.94 ± 0.80 \\
\bottomrule
\end{tabular}
\end{sc}
\end{small}
\end{center}
\vskip -0.1in
\end{table}

In summary, we demonstrated that using both reset and self-distillation can improve the performance of the VGG model on CIFAR-100. Furthermore, selecting the teacher from the training trajectory has been shown to be beneficial. However, there were limitations in selecting specific algorithms for the efficient operation of DFL. Consequently, in some cases, we observed that the self-distillation loss did not decrease during training, and DFL did not lead to improvements in the model’s performance.

\section{Conclusion}
\label{sec:con}
In this paper, we propose Diverse Feature Learning (DFL), which combines self-distillation and reset based on the necessity of feature preservation and new feature learning. We demonstrate that our proposed DFL can enhance performance, surpassing existing works in specific cases. It shows that using reset in conjunction with self-distillation exhibits a synergistic effect. Additionally, it has been shown that choosing multiple teachers appropriately on the training trajectory can be beneficial for self-distillation.

Our limitation lies in the fact that detailed algorithms have certain constraints when materializing the concepts of DFL. Specifically, using the previous epoch's training accuracy as a measure of meaningfulness for the teachers' update introduces a vulnerability to overfitting. This issue could be addressed by adopting the uncertainty measurement used in active learning \cite{takezoe2023deep}, as well as by assessing similarity between weights. Moreover, an analysis is required regarding the selection of layers most suitable for feature learning. Additionally, we could apply reset more stably \cite{anonymous2024drm, lee2023plastic}, and obtain model weights for self-distillation in various ways \cite{arani2022learning, tarvainen2017mean}.

\newpage
\section{Potential Broader Impact}
This paper presents work whose goal is to advance the field of Machine Learning. There are many potential societal consequences of our work, none which we feel must be specifically highlighted here.


\begin{thebibliography}{41}
\providecommand{\natexlab}[1]{#1}
\providecommand{\url}[1]{\texttt{#1}}
\expandafter\ifx\csname urlstyle\endcsname\relax
  \providecommand{\doi}[1]{doi: #1}\else
  \providecommand{\doi}{doi: \begingroup \urlstyle{rm}\Url}\fi

\bibitem[Allen-Zhu \& Li(2023)Allen-Zhu and Li]{allen-zhu2023towards}
Allen-Zhu, Z. and Li, Y.
\newblock Towards understanding ensemble, knowledge distillation and self-distillation in deep learning.
\newblock In \emph{The Eleventh International Conference on Learning Representations}, 2023.
\newblock URL \url{https://openreview.net/forum?id=Uuf2q9TfXGA}.

\bibitem[Arani et~al.(2022)Arani, Sarfraz, and Zonooz]{arani2022learning}
Arani, E., Sarfraz, F., and Zonooz, B.
\newblock Learning fast, learning slow: A general continual learning method based on complementary learning system.
\newblock In \emph{International Conference on Learning Representations}, 2022.
\newblock URL \url{https://openreview.net/forum?id=uxxFrDwrE7Y}.

\bibitem[Burda et~al.(2019)Burda, Edwards, Storkey, and Klimov]{burda2018exploration}
Burda, Y., Edwards, H., Storkey, A., and Klimov, O.
\newblock Exploration by random network distillation.
\newblock In \emph{International Conference on Learning Representations}, 2019.
\newblock URL \url{https://openreview.net/forum?id=H1lJJnR5Ym}.

\bibitem[Choi et~al.(2019)Choi, Kim, and Kim]{choi2019self}
Choi, J., Kim, T., and Kim, C.
\newblock Self-ensembling with gan-based data augmentation for domain adaptation in semantic segmentation.
\newblock In \emph{Proceedings of the IEEE/CVF International Conference on Computer Vision}, pp.\  6830--6840, 2019.

\bibitem[Choi et~al.(2022)Choi, Lee, Shin, Kim, Kim, and Yoon]{choi2022perception}
Choi, J., Lee, J., Shin, C., Kim, S., Kim, H., and Yoon, S.
\newblock Perception prioritized training of diffusion models.
\newblock In \emph{Proceedings of the IEEE/CVF Conference on Computer Vision and Pattern Recognition}, pp.\  11472--11481, 2022.

\bibitem[DeVries \& Taylor(2017)DeVries and Taylor]{devries2017improved}
DeVries, T. and Taylor, G.~W.
\newblock Improved regularization of convolutional neural networks with cutout.
\newblock \emph{arXiv preprint arXiv:1708.04552}, 2017.

\bibitem[Ganaie et~al.(2022)Ganaie, Hu, Malik, Tanveer, and Suganthan]{ganaie2022ensemble}
Ganaie, M.~A., Hu, M., Malik, A., Tanveer, M., and Suganthan, P.
\newblock Ensemble deep learning: A review.
\newblock \emph{Engineering Applications of Artificial Intelligence}, 115:\penalty0 105151, 2022.

\bibitem[Hicsonmez et~al.(2023)Hicsonmez, Samet, Akbas, and Duygulu]{hicsonmez2023improving}
Hicsonmez, S., Samet, N., Akbas, E., and Duygulu, P.
\newblock Improving sketch colorization using adversarial segmentation consistency.
\newblock \emph{arXiv preprint arXiv:2301.08590}, 2023.

\bibitem[Iandola et~al.(2016)Iandola, Han, Moskewicz, Ashraf, Dally, and Keutzer]{iandola2016squeezenet}
Iandola, F.~N., Han, S., Moskewicz, M.~W., Ashraf, K., Dally, W.~J., and Keutzer, K.
\newblock Squeezenet: Alexnet-level accuracy with 50x fewer parameters and $<$ 0.5 mb model size.
\newblock \emph{arXiv preprint arXiv:1602.07360}, 2016.

\bibitem[Jung et~al.(2023)Jung, Lee, Kim, Shaban, Boots, and Choo]{jung2023cafa}
Jung, S., Lee, J., Kim, N., Shaban, A., Boots, B., and Choo, J.
\newblock Cafa: Class-aware feature alignment for test-time adaptation.
\newblock In \emph{Proceedings of the IEEE/CVF International Conference on Computer Vision}, pp.\  19060--19071, 2023.

\bibitem[Ke et~al.(2022)Ke, Shao, Lin, Konishi, Kim, and Liu]{ke2022continual}
Ke, Z., Shao, Y., Lin, H., Konishi, T., Kim, G., and Liu, B.
\newblock Continual pre-training of language models.
\newblock In \emph{The Eleventh International Conference on Learning Representations}, 2022.

\bibitem[Kim et~al.(2021)Kim, Ji, Yoon, and Hwang]{kim2021self}
Kim, K., Ji, B., Yoon, D., and Hwang, S.
\newblock Self-knowledge distillation with progressive refinement of targets.
\newblock In \emph{Proceedings of the IEEE/CVF International Conference on Computer Vision}, pp.\  6567--6576, 2021.

\bibitem[Kim et~al.(2022)Kim, Hwang, Ahn, Park, and Kwak]{kim2022learning}
Kim, N., Hwang, S., Ahn, S., Park, J., and Kwak, S.
\newblock Learning debiased classifier with biased committee.
\newblock \emph{Advances in Neural Information Processing Systems}, 35:\penalty0 18403--18415, 2022.

\bibitem[Kirichenko et~al.(2023)Kirichenko, Izmailov, and Wilson]{kirichenko2023last}
Kirichenko, P., Izmailov, P., and Wilson, A.~G.
\newblock Last layer re-training is sufficient for robustness to spurious correlations.
\newblock In \emph{The Eleventh International Conference on Learning Representations}, 2023.
\newblock URL \url{https://openreview.net/forum?id=Zb6c8A-Fghk}.

\bibitem[Krizhevsky et~al.(2009)Krizhevsky, Hinton, et~al.]{krizhevsky2009learning}
Krizhevsky, A., Hinton, G., et~al.
\newblock Learning multiple layers of features from tiny images.
\newblock 2009.

\bibitem[Lee et~al.(2023)Lee, Cho, Kim, Gwak, Kim, Choo, Yun, and Yun]{lee2023plastic}
Lee, H., Cho, H., Kim, H., Gwak, D., Kim, J., Choo, J., Yun, S.-Y., and Yun, C.
\newblock Plastic: Improving input and label plasticity for sample efficient reinforcement learning.
\newblock In \emph{Thirty-seventh Conference on Neural Information Processing Systems}, 2023.

\bibitem[Li \& Vasconcelos(2020)Li and Vasconcelos]{li2020background}
Li, Y. and Vasconcelos, N.
\newblock Background data resampling for outlier-aware classification.
\newblock In \emph{Proceedings of the IEEE/CVF Conference on Computer Vision and Pattern Recognition}, pp.\  13218--13227, 2020.

\bibitem[Montero et~al.(2020)Montero, Ludwig, Costa, Malhotra, and Bowers]{montero2020role}
Montero, M.~L., Ludwig, C.~J., Costa, R.~P., Malhotra, G., and Bowers, J.
\newblock The role of disentanglement in generalisation.
\newblock In \emph{International Conference on Learning Representations}, 2020.

\bibitem[Nicolicioiu et~al.(2023)Nicolicioiu, Nicolicioiu, Alexe, and Teney]{nicolicioiu2023learning}
Nicolicioiu, A.~M., Nicolicioiu, A.~L., Alexe, B., and Teney, D.
\newblock Learning diverse features in vision transformers for improved generalization, 2023.

\bibitem[Nikishin et~al.(2023)Nikishin, Oh, Ostrovski, Lyle, Pascanu, Dabney, and Barreto]{nikishin2023deep}
Nikishin, E., Oh, J., Ostrovski, G., Lyle, C., Pascanu, R., Dabney, W., and Barreto, A.
\newblock Deep reinforcement learning with plasticity injection.
\newblock In \emph{Workshop on Reincarnating Reinforcement Learning at ICLR 2023}, 2023.
\newblock URL \url{https://openreview.net/forum?id=O9cJADBZT1}.

\bibitem[Park et~al.(2023)Park, Lee, Lee, and Ye]{park2023training}
Park, G.~Y., Lee, S., Lee, S.~W., and Ye, J.~C.
\newblock Training debiased subnetworks with contrastive weight pruning.
\newblock In \emph{Proceedings of the IEEE/CVF Conference on Computer Vision and Pattern Recognition}, pp.\  7929--7938, 2023.

\bibitem[Perera \& Patel(2023)Perera and Patel]{perera2023analyzing}
Perera, M.~V. and Patel, V.~M.
\newblock Analyzing bias in diffusion-based face generation models.
\newblock \emph{arXiv preprint arXiv:2305.06402}, 2023.

\bibitem[Sandler et~al.(2018)Sandler, Howard, Zhu, Zhmoginov, and Chen]{sandler2018mobilenetv2}
Sandler, M., Howard, A., Zhu, M., Zhmoginov, A., and Chen, L.-C.
\newblock Mobilenetv2: Inverted residuals and linear bottlenecks.
\newblock In \emph{Proceedings of the IEEE Conference on Computer Vision and Pattern Recognition}, pp.\  4510--4520, 2018.

\bibitem[Sauer et~al.(2021)Sauer, Chitta, M{\"u}ller, and Geiger]{sauer2021projected}
Sauer, A., Chitta, K., M{\"u}ller, J., and Geiger, A.
\newblock Projected gans converge faster.
\newblock \emph{Advances in Neural Information Processing Systems}, 34:\penalty0 17480--17492, 2021.

\bibitem[Schramowski et~al.(2023)Schramowski, Brack, Deiseroth, and Kersting]{schramowski2023safe}
Schramowski, P., Brack, M., Deiseroth, B., and Kersting, K.
\newblock Safe latent diffusion: Mitigating inappropriate degeneration in diffusion models.
\newblock In \emph{Proceedings of the IEEE/CVF Conference on Computer Vision and Pattern Recognition}, pp.\  22522--22531, 2023.

\bibitem[Simonyan \& Zisserman(2015)Simonyan and Zisserman]{simonyan2015very}
Simonyan, K. and Zisserman, A.
\newblock Very deep convolutional networks for large-scale image recognition.
\newblock In \emph{International Conference on Learning Representations}. Computational and Biological Learning Society, 2015.

\bibitem[Szegedy et~al.(2015)Szegedy, Liu, Jia, Sermanet, Reed, Anguelov, Erhan, Vanhoucke, and Rabinovich]{szegedy2015going}
Szegedy, C., Liu, W., Jia, Y., Sermanet, P., Reed, S., Anguelov, D., Erhan, D., Vanhoucke, V., and Rabinovich, A.
\newblock Going deeper with convolutions.
\newblock In \emph{Proceedings of the IEEE Conference on Computer Vision and Pattern Recognition}, pp.\  1--9, 2015.

\bibitem[Takezoe et~al.(2023)Takezoe, Liu, Mao, Chen, Feng, Zhang, Wang, et~al.]{takezoe2023deep}
Takezoe, R., Liu, X., Mao, S., Chen, M.~T., Feng, Z., Zhang, S., Wang, X., et~al.
\newblock Deep active learning for computer vision: Past and future.
\newblock \emph{APSIPA Transactions on Signal and Information Processing}, 12\penalty0 (1), 2023.

\bibitem[Tarvainen \& Valpola(2017)Tarvainen and Valpola]{tarvainen2017mean}
Tarvainen, A. and Valpola, H.
\newblock Mean teachers are better role models: Weight-averaged consistency targets improve semi-supervised deep learning results.
\newblock \emph{Advances in Neural Information Processing Systems}, 30, 2017.

\bibitem[Teterwak et~al.(2023)Teterwak, Saito, Tsiligkaridis, Saenko, and Plummer]{teterwak2023erm++}
Teterwak, P., Saito, K., Tsiligkaridis, T., Saenko, K., and Plummer, B.~A.
\newblock Erm++: An improved baseline for domain generalization.
\newblock \emph{ICML Workshop}, 2023.

\bibitem[Tseng et~al.(2020)Tseng, Lee, Huang, and Yang]{Tseng2020Cross-Domain}
Tseng, H.-Y., Lee, H.-Y., Huang, J.-B., and Yang, M.-H.
\newblock Cross-domain few-shot classification via learned feature-wise transformation.
\newblock In \emph{International Conference on Learning Representations}, 2020.
\newblock URL \url{https://openreview.net/forum?id=SJl5Np4tPr}.

\bibitem[Wang et~al.(2022)Wang, Fink, Van~Gool, and Dai]{Wang_2022_CVPR}
Wang, Q., Fink, O., Van~Gool, L., and Dai, D.
\newblock Continual test-time domain adaptation.
\newblock In \emph{Proceedings of the IEEE/CVF Conference on Computer Vision and Pattern Recognition (CVPR)}, pp.\  7201--7211, June 2022.

\bibitem[Weiaicunzai(2020)]{weiaicunzai_pytorch_cifar100}
Weiaicunzai.
\newblock Pytorch implementation of cifar-100 training.
\newblock \url{https://github.com/weiaicunzai/pytorch-cifar100}, 2020.
\newblock Accessed: 2023-12-01.

\bibitem[Wen et~al.(2021)Wen, Xu, Tan, Wu, and Wu]{wen2021debiased}
Wen, Z., Xu, G., Tan, M., Wu, Q., and Wu, Q.
\newblock Debiased visual question answering from feature and sample perspectives.
\newblock \emph{Advances in Neural Information Processing Systems}, 34:\penalty0 3784--3796, 2021.

\bibitem[Wortsman et~al.(2022)Wortsman, Ilharco, Gadre, Roelofs, Gontijo-Lopes, Morcos, Namkoong, Farhadi, Carmon, Kornblith, et~al.]{wortsman2022model}
Wortsman, M., Ilharco, G., Gadre, S.~Y., Roelofs, R., Gontijo-Lopes, R., Morcos, A.~S., Namkoong, H., Farhadi, A., Carmon, Y., Kornblith, S., et~al.
\newblock Model soups: averaging weights of multiple fine-tuned models improves accuracy without increasing inference time.
\newblock In \emph{International Conference on Machine Learning}, pp.\  23965--23998. PMLR, 2022.

\bibitem[Xu et~al.(2024)Xu, Zheng, Liang, Wang, Yuan, Ji, Luo, Liu, Yuan, Hua, Li, Ze, au2, Huang, and Xu]{anonymous2024drm}
Xu, G., Zheng, R., Liang, Y., Wang, X., Yuan, Z., Ji, T., Luo, Y., Liu, X., Yuan, J., Hua, P., Li, S., Ze, Y., au2, H. D.~I., Huang, F., and Xu, H.
\newblock Drm: Mastering visual reinforcement learning through dormant ratio minimization.
\newblock In \emph{The Twelfth International Conference on Learning Representations}, 2024.
\newblock URL \url{https://openreview.net/forum?id=MSe8YFbhUE}.

\bibitem[Yang et~al.(2022)Yang, Peng, Du, Tao, Tang, and Pan]{yang2022bistnet}
Yang, Y., Peng, Z., Du, X., Tao, Z., Tang, J., and Pan, J.
\newblock Bistnet: Semantic image prior guided bidirectional temporal feature fusion for deep exemplar-based video colorization.
\newblock \emph{arXiv preprint arXiv:2212.02268}, 2022.

\bibitem[Zhang et~al.(2021)Zhang, Bengio, Hardt, Recht, and Vinyals]{zhang2021understanding}
Zhang, C., Bengio, S., Hardt, M., Recht, B., and Vinyals, O.
\newblock Understanding deep learning (still) requires rethinking generalization.
\newblock \emph{Communications of the ACM}, 64\penalty0 (3):\penalty0 107--115, 2021.

\bibitem[Zhang et~al.(2019)Zhang, Song, Gao, Chen, Bao, and Ma]{zhang2019your}
Zhang, L., Song, J., Gao, A., Chen, J., Bao, C., and Ma, K.
\newblock Be your own teacher: Improve the performance of convolutional neural networks via self distillation.
\newblock In \emph{Proceedings of the IEEE/CVF International Conference on Computer Vision}, pp.\  3713--3722, 2019.

\bibitem[Zhang et~al.(2018)Zhang, Zhou, Lin, and Sun]{zhang2018shufflenet}
Zhang, X., Zhou, X., Lin, M., and Sun, J.
\newblock Shufflenet: An extremely efficient convolutional neural network for mobile devices.
\newblock In \emph{Proceedings of the IEEE Conference on Computer Vision and Pattern Recognition}, pp.\  6848--6856, 2018.

\bibitem[Zhou et~al.(2022)Zhou, Vani, Larochelle, and Courville]{zhou2022fortuitous}
Zhou, H., Vani, A., Larochelle, H., and Courville, A.
\newblock Fortuitous forgetting in connectionist networks.
\newblock In \emph{International Conference on Learning Representations}, 2022.
\newblock URL \url{https://openreview.net/forum?id=ei3SY1_zYsE}.

\end{thebibliography}

\newpage
\appendix
\onecolumn
\section{Overall Experiment Results}
\label{app}
The overall experimental results using our Diverse Feature Learning (DFL) for CIFAR-100 and CIFAR-10 are presented in Table~\ref{tbl:cifar100} and Table~\ref{tbl:cifar10}, respectively. These tables record the test accuracy at the last step of training and include two notes. 

Firstly, despite being the same experiment, differences occur due to the use of different GPUs. Specifically, a GeForce 2080Ti and a TITAN RTX GPU were used, leading to minor differences in the results of VGG(1) and VGG(3), which vary in the number of layers for the student, when DFL is not applied. Therefore, in the main text tables, the mean and standard deviation are calculated and presented as a single mean and standard deviation using the formulas \(\bar{x} = \frac{\bar{x}_A + \bar{x}_B}{2}\) and \(s = \sqrt{\frac{s_A^2 + s_B^2}{2} + \frac{(\bar{x}_A - \bar{x})^2 + (\bar{x}_B - \bar{x})^2}{2}}\).

The details of the second note are as follows: CIFAR-100 and CIFAR-10 results in Tables~\ref{tbl:cifar100} and \ref{tbl:cifar10} include an ablation study where the last 10 of the 200 epochs were trained without DFL. The results for the last step of training, excluding these epochs, show the same characteristic with Tables~\ref{tbl:cifar100} and \ref{tbl:cifar10}, as demonstrated in Tables~\ref{tbl:cifar100_190} and \ref{tbl:cifar10_190}.

\begin{figure}[h!]
\vskip 0.2in
\begin{center}
\centerline{\includegraphics[width=0.9\columnwidth]{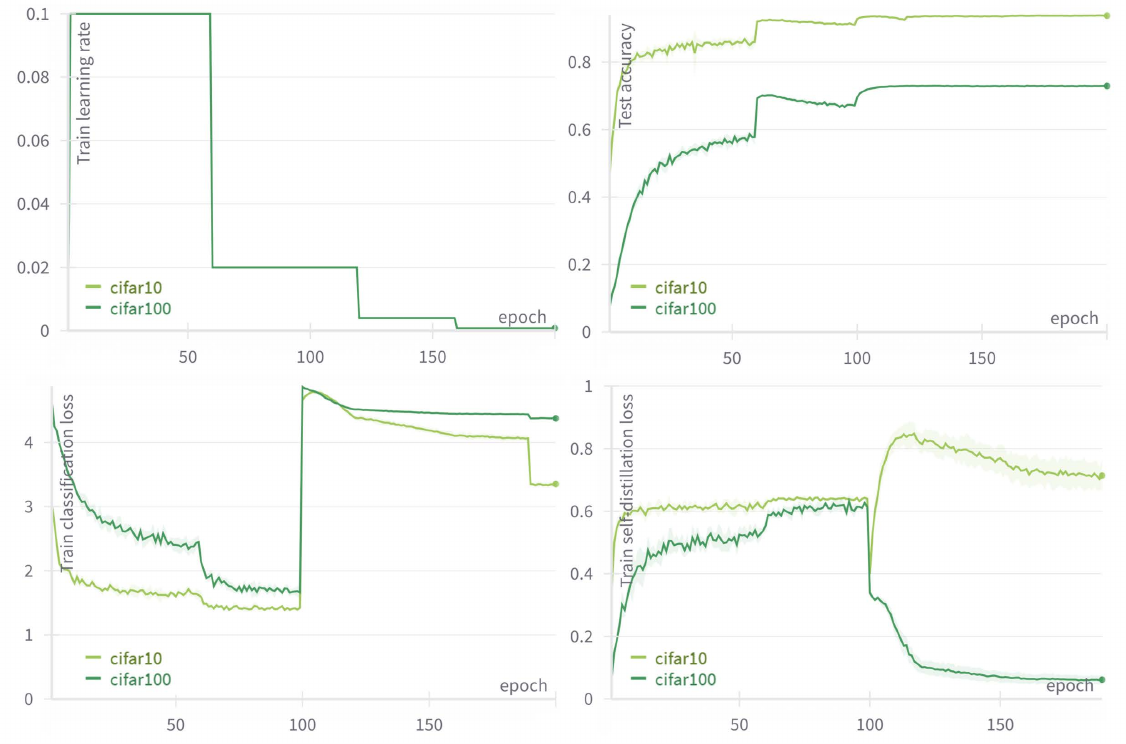}}
\caption{\textbf{Different Dataset} It represents the mean and standard deviation of the learning rate, test accuracy, training classification loss, and training self-distillation loss using the default settings in Table~\ref{tbl:ablation}. The lines in the graph represent the mean values, while the shaded areas indicate the standard deviations.}
\label{fig:default_dataset}
\end{center}
\vskip -0.2in
\end{figure}

Additionally, the learning process for the default setting in Table~\ref{tbl:ablation} is depicted through the graph in Figure~\ref{fig:default_dataset}, with all graphs drawn with respect to the training epoch. The top left graph shows the changes in the learning rate during the training process. The top right graph displays the changes in test accuracy as learning progresses. The bottom left graph illustrates the changes in classification loss during the training process. Lastly, the bottom right graph represents the changes in self-distillation loss during the training process.

\begin{table}[H]
\caption{\textbf{Test Accuracy on CIFAR-100} It represents the mean accuracy and standard deviation for five seeds across each of the five models. The content in parentheses indicates the number of last layers used for the reset; if not specified, one layer is assumed. The meanings of each abbreviation in the table's headline are as follows: KL refers to self-distillation. K represents the number of teachers, where a value of zero indicates that self-distillation is not utilized. Regarding the reset, R indicates whether it is used; thus, when R is zero, reset is not utilized. $T_{\text{cycle}}$ signifies the duration of the cycle used for the teacher update and the student reset, expressed in epochs. M indicates the re-initialization method used for reset, where a value of one means that re-initialization was done using the mean of the teachers' weights. A value of zero signifies that random weights were utilized for re-initialization.}
\label{tbl:cifar100}
\vskip 0.15in
\begin{center}
\begin{small}
\begin{sc}
\begin{tabular}{|c|c|c|c|c|c|c|c|c|c|}
\toprule
 \multicolumn{1}{|c|}{KL} & \multicolumn{3}{|c|}{RESET} & \multicolumn{6}{|c|}{Model} \\
\cmidrule(lr){1-1} \cmidrule(lr){2-4} \cmidrule(lr){5-10}
 K & R & $T_{\text{cycle}}$ & M & googlenet  & mobilenetv2 & shufflenet & squeezenet & vgg(3) & vgg(1) \\
\midrule
0 & 0 & 0   & 0 &  76.52 ± 0.39 &  70.30 ± 0.29 &  70.45 ± 0.35 &  69.53 ± 0.41 &   71.82 ± 0.34 &  71.88 ± 0.29 \\
  & 1 & 1   & 0 &  74.06 ± 0.27 &  68.60 ± 0.25 &  68.84 ± 0.37 &  61.29 ± 0.99 &  29.35 ± 16.11 &  71.81 ± 0.26 \\
  &   &     & 1 &  76.89 ± 0.22 &  70.05 ± 0.29 &  70.86 ± 0.40 &  68.99 ± 0.39 &   72.41 ± 0.23 &  71.83 ± 0.23 \\
  &   & 20  & 0 &  76.67 ± 0.25 &  69.65 ± 0.39 &  70.07 ± 0.14 &  68.59 ± 0.60 &   69.06 ± 2.50 &  71.72 ± 0.35 \\
  &   &     & 1 &  76.62 ± 0.32 &  69.62 ± 0.41 &  70.17 ± 0.23 &  68.49 ± 0.35 &   71.46 ± 0.27 &  71.77 ± 0.32 \\
  &   & 50  & 0 &  77.00 ± 0.43 &  69.82 ± 0.31 &  70.07 ± 0.49 &  69.20 ± 0.32 &   72.90 ± 0.33 &  71.90 ± 0.28 \\
  &   &     & 1 &  76.67 ± 0.18 &  69.76 ± 0.25 &  70.68 ± 0.38 &  69.51 ± 0.25 &   71.61 ± 0.14 &  71.97 ± 0.29 \\
  &   & 100 & 0 &  76.86 ± 0.28 &  69.73 ± 0.16 &  70.59 ± 0.39 &  69.58 ± 0.34 &   72.66 ± 0.20 &  71.95 ± 0.28 \\
  &   &     & 1 &  76.46 ± 0.15 &  69.91 ± 0.46 &  70.37 ± 0.36 &  69.80 ± 0.24 &   71.71 ± 0.28 &  71.95 ± 0.28 \\
1 & 0 & 0   & 0 &  76.10 ± 0.16 &  70.17 ± 0.20 &  70.94 ± 0.24 &  70.25 ± 0.23 &   72.18 ± 0.24 &  72.35 ± 0.18 \\
  & 1 & 1   & 0 &  73.98 ± 0.43 &  69.21 ± 0.35 &  69.06 ± 0.56 &  58.19 ± 1.28 &   39.30 ± 3.88 &  71.89 ± 0.22 \\
  &   &     & 1 &  77.08 ± 0.24 &  70.22 ± 0.33 &  70.89 ± 0.42 &  68.86 ± 0.25 &   72.18 ± 0.28 &  71.92 ± 0.21 \\
  &   & 20  & 0 &  75.50 ± 0.55 &  69.99 ± 0.23 &  70.18 ± 0.22 &  68.96 ± 0.19 &   70.22 ± 1.70 &  72.02 ± 0.33 \\
  &   &     & 1 &  75.95 ± 0.72 &  69.47 ± 0.61 &  70.60 ± 0.39 &  69.33 ± 0.42 &   71.71 ± 0.21 &  72.02 ± 0.33 \\
  &   & 50  & 0 &  76.06 ± 0.27 &  70.08 ± 0.23 &  70.54 ± 0.76 &  69.51 ± 0.28 &   72.97 ± 0.34 &  72.14 ± 0.20 \\
  &   &     & 1 &  75.89 ± 1.05 &  69.88 ± 0.27 &  71.18 ± 0.31 &  69.37 ± 0.38 &   72.12 ± 0.42 &  72.18 ± 0.17 \\
  &   & 100 & 0 &  75.10 ± 1.20 &  69.90 ± 0.48 &  70.29 ± 0.26 &  69.08 ± 0.35 &   72.75 ± 0.27 &  72.06 ± 0.22 \\
  &   &     & 1 &  74.99 ± 1.37 &  70.68 ± 0.61 &  71.67 ± 0.68 &  70.65 ± 0.27 &   72.89 ± 0.33 &  72.14 ± 0.22 \\
2 & 0 & 0   & 0 &  75.88 ± 1.07 &  70.38 ± 0.21 &  70.64 ± 0.47 &  70.52 ± 0.41 &   72.24 ± 0.17 &  72.37 ± 0.24 \\
  & 1 & 1   & 0 &  74.22 ± 0.22 &  68.99 ± 0.19 &  68.96 ± 0.33 &  60.52 ± 1.66 &   37.68 ± 3.76 &  71.78 ± 0.26 \\
  &   &     & 1 &  76.69 ± 0.28 &  69.97 ± 0.46 &  70.89 ± 0.29 &  68.88 ± 0.14 &   71.79 ± 0.37 &  71.72 ± 0.21 \\
  &   & 20  & 0 &  75.22 ± 0.51 &  70.16 ± 0.23 &  70.28 ± 0.37 &  68.76 ± 0.30 &   70.03 ± 3.50 &  72.08 ± 0.21 \\
  &   &     & 1 &  75.56 ± 1.21 &  70.12 ± 0.50 &  71.08 ± 0.47 &  69.57 ± 0.12 &   72.26 ± 0.18 &  72.01 ± 0.05 \\
  &   & 50  & 0 &  75.89 ± 0.57 &  70.21 ± 0.40 &  70.69 ± 0.41 &  69.67 ± 0.12 &   72.71 ± 0.14 &  71.95 ± 0.27 \\
  &   &     & 1 &  76.38 ± 0.47 &  70.29 ± 0.37 &  70.81 ± 0.73 &  69.91 ± 0.15 &   72.69 ± 0.35 &  71.96 ± 0.24 \\
  &   & 100 & 0 &  76.07 ± 0.29 &  69.97 ± 0.42 &  70.40 ± 0.25 &  69.42 ± 0.29 &   72.84 ± 0.18 &  72.39 ± 0.30 \\
  &   &     & 1 &  76.35 ± 0.51 &  70.57 ± 0.45 &  71.27 ± 0.32 &  70.17 ± 0.40 &   73.03 ± 0.24 &  72.26 ± 0.34 \\
4 & 0 & 0   & 0 &  75.71 ± 1.55 &  70.45 ± 0.24 &  70.94 ± 0.38 &  70.11 ± 0.22 &   72.21 ± 0.25 &  72.21 ± 0.29 \\
  & 1 & 1   & 0 &  74.25 ± 0.34 &  69.33 ± 0.10 &  69.20 ± 0.37 &  60.18 ± 0.54 &   40.62 ± 3.77 &  72.00 ± 0.35 \\
  &   &     & 1 &  76.20 ± 0.08 &  69.77 ± 0.40 &  70.58 ± 0.49 &  68.68 ± 0.33 &   72.10 ± 0.30 &  72.11 ± 0.24 \\
  &   & 20  & 0 &  75.43 ± 0.48 &  70.15 ± 0.47 &  70.06 ± 0.48 &  68.92 ± 0.36 &   70.70 ± 2.79 &  71.95 ± 0.23 \\
  &   &     & 1 &  76.47 ± 0.28 &  70.28 ± 0.16 &  71.04 ± 0.16 &  69.40 ± 0.48 &   71.63 ± 0.34 &  72.20 ± 0.17 \\
  &   & 50  & 0 &  74.83 ± 1.95 &  70.05 ± 0.42 &  70.25 ± 0.28 &  69.31 ± 0.40 &   72.68 ± 0.21 &  71.87 ± 0.39 \\
  &   &     & 1 &  75.62 ± 1.01 &  70.18 ± 0.14 &  71.18 ± 0.30 &  69.84 ± 0.25 &   72.56 ± 0.22 &  72.02 ± 0.32 \\
  &   & 100 & 0 &  76.35 ± 0.23 &  69.97 ± 0.36 &  70.69 ± 0.58 &  69.33 ± 0.21 &   73.04 ± 0.09 &  72.20 ± 0.26 \\
  &   &     & 1 &  75.94 ± 0.80 &  70.14 ± 0.43 &  71.18 ± 0.59 &  70.14 ± 0.22 &   72.94 ± 0.44 &  72.38 ± 0.18 \\
8 & 0 & 0   & 0 &  76.01 ± 0.40 &  70.55 ± 0.20 &  70.71 ± 0.19 &  70.18 ± 0.11 &   72.26 ± 0.28 &  72.42 ± 0.17 \\
  & 1 & 1   & 0 &  74.47 ± 0.22 &  69.29 ± 0.48 &  69.14 ± 0.37 &  60.72 ± 0.67 &   40.76 ± 1.97 &  72.05 ± 0.39 \\
  &   &     & 1 &  75.47 ± 0.42 &  70.28 ± 0.14 &  70.08 ± 0.31 &  68.71 ± 0.50 &   72.01 ± 0.33 &  71.67 ± 0.36 \\
  &   & 20  & 0 &  75.05 ± 0.89 &  70.17 ± 0.16 &  70.30 ± 0.44 &  68.83 ± 0.38 &   69.28 ± 1.48 &  72.03 ± 0.09 \\
  &   &     & 1 &  75.65 ± 1.38 &  70.31 ± 0.37 &  70.55 ± 0.34 &  69.09 ± 0.26 &   69.43 ± 1.32 &  71.97 ± 0.30 \\
  &   & 50  & 0 &  75.34 ± 0.96 &  70.14 ± 0.36 &  70.33 ± 0.40 &  69.70 ± 0.09 &   72.79 ± 0.37 &  72.23 ± 0.28 \\
  &   &     & 1 &  75.83 ± 0.60 &  70.22 ± 0.25 &  70.75 ± 0.58 &  69.29 ± 0.22 &   68.05 ± 1.49 &  72.50 ± 0.38 \\
  &   & 100 & 0 &  76.32 ± 0.33 &  69.91 ± 0.45 &  70.29 ± 0.49 &  69.58 ± 0.17 &   73.00 ± 0.09 &  72.48 ± 0.32 \\
  &   &     & 1 &  75.45 ± 1.47 &  69.94 ± 0.37 &  71.06 ± 0.30 &  69.77 ± 0.54 &   68.00 ± 2.09 &  72.31 ± 0.31 \\
\bottomrule
\end{tabular}
\end{sc}
\end{small}
\end{center}
\vskip -0.1in
\end{table}

\begin{table}[H]
\caption{\textbf{Test Accuracy on CIFAR-10} It represents the mean accuracy and standard deviation for five seeds across each of the five models. The content in parentheses indicates the number of last layers used for the reset; if not specified, one layer is assumed. The meanings of each abbreviation in the table's headline are as follows: KL refers to self-distillation. K represents the number of teachers, where a value of zero indicates that self-distillation is not utilized. Regarding the reset, R indicates whether it is used; thus, when R is zero, reset is not utilized. $T_{\text{cycle}}$ signifies the duration of the cycle used for the teacher update and the student reset, expressed in epochs. M indicates the re-initialization method used for reset, where a value of one means that re-initialization was done using the mean of the teachers' weights. A value of zero signifies that random weights were utilized for re-initialization.}
\label{tbl:cifar10}
\vskip 0.15in
\begin{center}
\begin{small}
\begin{sc}
\begin{tabular}{|c|c|c|c|c|c|c|c|c|c|}
\toprule
 \multicolumn{1}{|c|}{KL} & \multicolumn{3}{|c|}{RESET} & \multicolumn{6}{|c|}{Model} \\
\cmidrule(lr){1-1} \cmidrule(lr){2-4} \cmidrule(lr){5-10}
 K & R & $T_{\text{cycle}}$ & M & googlenet  & mobilenetv2 & shufflenet & squeezenet & vgg(3) & vgg(1) \\
\midrule
0 & 0 & 0   & 0 &  94.62 ± 0.05 &  91.81 ± 0.25 &  92.21 ± 0.35 &  92.35 ± 0.13 &  93.68 ± 0.13 &  93.59 ± 0.13 \\
  & 1 & 1   & 0 &  94.27 ± 0.19 &  91.52 ± 0.21 &  91.75 ± 0.26 &  91.96 ± 0.19 &  93.52 ± 0.20 &  93.66 ± 0.19 \\
  &   &     & 1 &  94.59 ± 0.14 &  91.74 ± 0.09 &  92.30 ± 0.30 &  92.41 ± 0.15 &  93.65 ± 0.21 &  93.61 ± 0.21 \\
  &   & 20  & 0 &  94.35 ± 0.24 &  91.81 ± 0.09 &  92.11 ± 0.16 &  92.34 ± 0.21 &  93.59 ± 0.06 &  93.51 ± 0.09 \\
  &   &     & 1 &  94.66 ± 0.16 &  91.95 ± 0.15 &  92.26 ± 0.16 &  92.17 ± 0.18 &  93.64 ± 0.21 &  93.57 ± 0.09 \\
  &   & 50  & 0 &  94.54 ± 0.15 &  91.85 ± 0.18 &  92.38 ± 0.17 &  92.59 ± 0.17 &  93.63 ± 0.13 &  93.54 ± 0.27 \\
  &   &     & 1 &  94.58 ± 0.11 &  91.86 ± 0.20 &  92.49 ± 0.43 &  92.18 ± 0.11 &  93.68 ± 0.29 &  93.49 ± 0.18 \\
  &   & 100 & 0 &  94.62 ± 0.23 &  91.68 ± 0.13 &  92.33 ± 0.17 &  92.28 ± 0.13 &  93.73 ± 0.08 &  93.59 ± 0.12 \\
  &   &     & 1 &  94.66 ± 0.20 &  91.86 ± 0.14 &  92.29 ± 0.25 &  92.25 ± 0.23 &  93.68 ± 0.18 &  93.59 ± 0.12 \\
1 & 0 & 0   & 0 &  94.42 ± 0.19 &  91.63 ± 0.11 &  92.34 ± 0.27 &  92.26 ± 0.19 &  93.53 ± 0.08 &  93.45 ± 0.20 \\
  & 1 & 1   & 0 &  94.19 ± 0.33 &  91.56 ± 0.21 &  91.97 ± 0.07 &  91.92 ± 0.18 &  93.33 ± 0.13 &  93.71 ± 0.18 \\
  &   &     & 1 &  94.50 ± 0.18 &  91.85 ± 0.29 &  92.29 ± 0.20 &  92.62 ± 0.27 &  93.55 ± 0.12 &  93.71 ± 0.18 \\
  &   & 20  & 0 &  94.36 ± 0.14 &  91.54 ± 0.14 &  91.98 ± 0.25 &  92.20 ± 0.16 &  93.80 ± 0.11 &  93.68 ± 0.21 \\
  &   &     & 1 &  94.68 ± 0.12 &  91.71 ± 0.14 &  92.30 ± 0.32 &  92.30 ± 0.15 &  93.68 ± 0.10 &  93.70 ± 0.18 \\
  &   & 50  & 0 &  94.40 ± 0.10 &  91.63 ± 0.15 &  92.31 ± 0.17 &  92.41 ± 0.19 &  93.64 ± 0.20 &  93.67 ± 0.13 \\
  &   &     & 1 &  94.25 ± 0.37 &  91.76 ± 0.11 &  92.20 ± 0.17 &  92.43 ± 0.19 &  93.71 ± 0.12 &  93.67 ± 0.13 \\
  &   & 100 & 0 &  94.45 ± 0.15 &  91.55 ± 0.12 &  92.58 ± 0.25 &  92.36 ± 0.11 &  93.63 ± 0.18 &  93.65 ± 0.06 \\
  &   &     & 1 &  94.48 ± 0.22 &  91.80 ± 0.05 &  92.32 ± 0.33 &  92.43 ± 0.14 &  93.71 ± 0.12 &  93.65 ± 0.06 \\
2 & 0 & 0   & 0 &  94.52 ± 0.13 &  91.83 ± 0.18 &  92.22 ± 0.22 &  92.23 ± 0.27 &  93.53 ± 0.12 &  93.42 ± 0.16 \\
  & 1 & 1   & 0 &  94.26 ± 0.18 &  91.85 ± 0.19 &  91.78 ± 0.23 &  91.77 ± 0.30 &  93.30 ± 0.23 &  93.70 ± 0.16 \\
  &   &     & 1 &  94.46 ± 0.15 &  91.79 ± 0.16 &  92.03 ± 0.26 &  92.69 ± 0.12 &  93.65 ± 0.20 &  93.59 ± 0.21 \\
  &   & 20  & 0 &  94.17 ± 0.43 &  91.65 ± 0.19 &  92.32 ± 0.12 &  92.34 ± 0.10 &  93.65 ± 0.18 &  93.47 ± 0.16 \\
  &   &     & 1 &  94.60 ± 0.08 &  91.86 ± 0.17 &  92.27 ± 0.15 &  92.42 ± 0.22 &  93.62 ± 0.25 &  93.63 ± 0.15 \\
  &   & 50  & 0 &  94.33 ± 0.16 &  91.65 ± 0.31 &  92.31 ± 0.19 &  92.48 ± 0.33 &  93.52 ± 0.21 &  93.56 ± 0.12 \\
  &   &     & 1 &  94.36 ± 0.18 &  91.92 ± 0.23 &  92.32 ± 0.26 &  92.47 ± 0.23 &  93.64 ± 0.24 &  93.56 ± 0.07 \\
  &   & 100 & 0 &  94.23 ± 0.41 &  91.62 ± 0.21 &  92.58 ± 0.09 &  92.32 ± 0.19 &  93.70 ± 0.13 &  93.44 ± 0.23 \\
  &   &     & 1 &  94.45 ± 0.11 &  91.82 ± 0.20 &  92.19 ± 0.24 &  92.50 ± 0.21 &  93.77 ± 0.18 &  93.47 ± 0.18 \\
4 & 0 & 0   & 0 &  94.39 ± 0.17 &  91.92 ± 0.15 &  92.32 ± 0.24 &  92.34 ± 0.22 &  93.45 ± 0.11 &  93.50 ± 0.31 \\
  & 1 & 1   & 0 &  94.22 ± 0.08 &  91.70 ± 0.19 &  91.99 ± 0.22 &  91.97 ± 0.21 &  93.43 ± 0.14 &  93.55 ± 0.08 \\
  &   &     & 1 &  94.46 ± 0.24 &  91.93 ± 0.29 &  91.96 ± 0.19 &  92.36 ± 0.16 &  93.53 ± 0.12 &  93.55 ± 0.12 \\
  &   & 20  & 0 &  94.32 ± 0.17 &  91.53 ± 0.15 &  92.23 ± 0.24 &  92.32 ± 0.22 &  93.70 ± 0.30 &  93.73 ± 0.07 \\
  &   &     & 1 &  94.43 ± 0.13 &  91.67 ± 0.25 &  92.15 ± 0.19 &  92.64 ± 0.16 &  93.48 ± 0.15 &  93.59 ± 0.18 \\
  &   & 50  & 0 &  94.35 ± 0.17 &  91.73 ± 0.27 &  92.30 ± 0.16 &  92.53 ± 0.13 &  93.63 ± 0.24 &  93.61 ± 0.14 \\
  &   &     & 1 &  94.51 ± 0.11 &  91.86 ± 0.16 &  92.46 ± 0.24 &  92.44 ± 0.08 &  93.61 ± 0.06 &  93.54 ± 0.10 \\
  &   & 100 & 0 &  94.42 ± 0.22 &  91.58 ± 0.29 &  92.43 ± 0.20 &  92.48 ± 0.14 &  93.64 ± 0.14 &  93.59 ± 0.20 \\
  &   &     & 1 &  94.48 ± 0.14 &  91.77 ± 0.21 &  92.60 ± 0.14 &  92.50 ± 0.10 &  93.78 ± 0.10 &  93.49 ± 0.31 \\
8 & 0 & 0   & 0 &  94.57 ± 0.12 &  91.85 ± 0.26 &  92.25 ± 0.15 &  92.42 ± 0.12 &  93.48 ± 0.18 &  93.52 ± 0.14 \\
  & 1 & 1   & 0 &  94.20 ± 0.13 &  91.78 ± 0.13 &  92.03 ± 0.22 &  91.75 ± 0.29 &  93.27 ± 0.16 &  93.78 ± 0.16 \\
  &   &     & 1 &  94.39 ± 0.30 &  91.82 ± 0.17 &  91.66 ± 0.27 &  92.43 ± 0.02 &  93.62 ± 0.18 &  93.74 ± 0.16 \\
  &   & 20  & 0 &  94.27 ± 0.34 &  91.68 ± 0.11 &  92.41 ± 0.20 &  92.44 ± 0.19 &  93.57 ± 0.06 &  93.69 ± 0.09 \\
  &   &     & 1 &  94.50 ± 0.08 &  91.70 ± 0.15 &  92.20 ± 0.20 &  92.49 ± 0.30 &  93.14 ± 0.23 &  93.77 ± 0.19 \\
  &   & 50  & 0 &  94.29 ± 0.13 &  91.78 ± 0.21 &  92.27 ± 0.27 &  92.46 ± 0.09 &  93.67 ± 0.12 &  93.58 ± 0.13 \\
  &   &     & 1 &  94.42 ± 0.19 &  91.83 ± 0.11 &  92.29 ± 0.14 &  92.45 ± 0.27 &  93.08 ± 0.57 &  93.66 ± 0.15 \\
  &   & 100 & 0 &  94.37 ± 0.22 &  91.51 ± 0.19 &  92.35 ± 0.21 &  92.34 ± 0.18 &  93.62 ± 0.19 &  93.57 ± 0.10 \\
  &   &     & 1 &  94.37 ± 0.17 &  91.55 ± 0.27 &  92.41 ± 0.28 &  92.52 ± 0.12 &  93.22 ± 0.13 &  93.59 ± 0.11 \\
\bottomrule
\end{tabular}
\end{sc}
\end{small}
\end{center}
\vskip -0.1in
\end{table}

\begin{table}[H]
\caption{\textbf{Test Accuracy on CIFAR-100 over 190 Epochs} It represents the mean accuracy and standard deviation for five seeds across each of the five models. The content in parentheses indicates the number of last layers used for the reset; if not specified, one layer is assumed. The meanings of each abbreviation in the table's headline are as follows: KL refers to self-distillation. K represents the number of teachers, where a value of zero indicates that self-distillation is not utilized. Regarding the reset, R indicates whether it is used; thus, when R is zero, reset is not utilized. $T_{\text{cycle}}$ signifies the duration of the cycle used for the teacher update and the student reset, expressed in epochs. M indicates the re-initialization method used for reset, where a value of one means that re-initialization was done using the mean of the teachers' weights. A value of zero signifies that random weights were utilized for re-initialization.}
\label{tbl:cifar100_190}
\vskip 0.15in
\begin{center}
\begin{small}
\begin{sc}
\begin{tabular}{|c|c|c|c|c|c|c|c|c|c|}
\toprule
 \multicolumn{1}{|c|}{KL} & \multicolumn{3}{|c|}{RESET} & \multicolumn{6}{|c|}{Model} \\
\cmidrule(lr){1-1} \cmidrule(lr){2-4} \cmidrule(lr){5-10}
K & R & $T_{\text{cycle}}$ & M & googlenet  & mobilenetv2 & shufflenet & squeezenet & vgg(3) & vgg(1) \\
\midrule
0 & 0 & 0   & 0 &  76.52 ± 0.28 &  70.19 ± 0.36 &  70.48 ± 0.39 &  69.46 ± 0.33 &  71.73 ± 0.28 &  72.01 ± 0.19 \\
  & 1 & 1   & 0 &  73.30 ± 0.80 &  68.35 ± 0.31 &  66.51 ± 0.29 &  41.27 ± 2.25 &  15.89 ± 8.86 &  71.81 ± 0.31 \\
  &   &     & 1 &  76.85 ± 0.29 &  70.20 ± 0.24 &  70.93 ± 0.32 &  68.99 ± 0.45 &  72.38 ± 0.28 &  71.82 ± 0.27 \\
  &   & 20  & 0 &  76.66 ± 0.26 &  69.71 ± 0.22 &  70.04 ± 0.29 &  68.26 ± 0.79 &  68.31 ± 2.75 &  71.77 ± 0.38 \\
  &   &     & 1 &  76.59 ± 0.20 &  69.56 ± 0.33 &  70.14 ± 0.24 &  68.64 ± 0.41 &  71.62 ± 0.29 &  71.82 ± 0.35 \\
  &   & 50  & 0 &  76.93 ± 0.37 &  69.92 ± 0.23 &  70.23 ± 0.44 &  69.42 ± 0.38 &  72.96 ± 0.24 &  71.90 ± 0.09 \\
  &   &     & 1 &  76.60 ± 0.17 &  69.91 ± 0.29 &  70.69 ± 0.35 &  69.63 ± 0.26 &  71.66 ± 0.20 &  71.97 ± 0.18 \\
  &   & 100 & 0 &  76.79 ± 0.22 &  69.84 ± 0.23 &  70.65 ± 0.33 &  69.39 ± 0.47 &  72.82 ± 0.05 &  71.83 ± 0.22 \\
  &   &     & 1 &  76.45 ± 0.17 &  69.96 ± 0.46 &  70.44 ± 0.38 &  69.86 ± 0.18 &  71.72 ± 0.32 &  71.83 ± 0.22 \\
1 & 0 & 0   & 0 &  76.30 ± 0.24 &  70.74 ± 0.29 &  71.31 ± 0.33 &  70.37 ± 0.31 &  72.23 ± 0.31 &  72.45 ± 0.18 \\
  & 1 & 1   & 0 &  73.61 ± 0.39 &  67.03 ± 2.21 &  63.67 ± 1.57 &  37.06 ± 5.21 &  21.11 ± 3.54 &  71.83 ± 0.16 \\
  &   &     & 1 &  77.08 ± 0.36 &  70.24 ± 0.48 &  70.90 ± 0.46 &  68.89 ± 0.42 &  72.05 ± 0.35 &  71.94 ± 0.14 \\
  &   & 20  & 0 &  75.43 ± 0.54 &  70.50 ± 0.42 &  70.45 ± 0.28 &  68.58 ± 0.23 &  68.67 ± 1.68 &  71.94 ± 0.42 \\
  &   &     & 1 &  75.89 ± 0.80 &  69.50 ± 0.61 &  70.53 ± 0.38 &  69.37 ± 0.36 &  71.81 ± 0.22 &  71.94 ± 0.42 \\
  &   & 50  & 0 &  76.03 ± 0.28 &  70.41 ± 0.27 &  70.87 ± 0.73 &  69.39 ± 0.29 &  72.86 ± 0.39 &  72.03 ± 0.44 \\
  &   &     & 1 &  75.83 ± 1.03 &  69.84 ± 0.38 &  71.06 ± 0.42 &  69.29 ± 0.40 &  72.13 ± 0.19 &  72.11 ± 0.43 \\
  &   & 100 & 0 &  75.16 ± 1.21 &  70.51 ± 0.40 &  70.36 ± 0.24 &  67.97 ± 0.40 &  72.62 ± 0.11 &  72.06 ± 0.29 \\
  &   &     & 1 &  75.03 ± 1.40 &  70.98 ± 0.45 &  71.95 ± 0.75 &  70.59 ± 0.28 &  72.41 ± 0.35 &  72.07 ± 0.30 \\
2 & 0 & 0   & 0 &  75.86 ± 1.12 &  70.88 ± 0.18 &  71.11 ± 0.47 &  70.54 ± 0.34 &  72.37 ± 0.18 &  72.50 ± 0.25 \\
  & 1 & 1   & 0 &  73.98 ± 0.28 &  66.85 ± 1.14 &  64.74 ± 1.03 &  42.51 ± 2.54 &  17.18 ± 5.40 &  71.86 ± 0.23 \\
  &   &     & 1 &  76.82 ± 0.27 &  70.04 ± 0.26 &  70.98 ± 0.11 &  68.79 ± 0.31 &  71.81 ± 0.27 &  71.77 ± 0.39 \\
  &   & 20  & 0 &  75.13 ± 0.26 &  70.51 ± 0.26 &  70.48 ± 0.33 &  68.35 ± 0.26 &  69.46 ± 4.21 &  72.11 ± 0.22 \\
  &   &     & 1 &  75.57 ± 1.28 &  70.25 ± 0.39 &  71.06 ± 0.25 &  69.55 ± 0.23 &  72.16 ± 0.21 &  72.09 ± 0.14 \\
  &   & 50  & 0 &  75.96 ± 0.50 &  70.71 ± 0.36 &  70.92 ± 0.50 &  69.55 ± 0.18 &  72.82 ± 0.10 &  72.00 ± 0.30 \\
  &   &     & 1 &  76.04 ± 0.39 &  70.38 ± 0.36 &  70.70 ± 0.69 &  70.02 ± 0.05 &  72.67 ± 0.23 &  71.96 ± 0.17 \\
  &   & 100 & 0 &  76.08 ± 0.33 &  70.62 ± 0.25 &  70.63 ± 0.22 &  69.30 ± 0.35 &  72.87 ± 0.11 &  72.38 ± 0.22 \\
  &   &     & 1 &  76.27 ± 0.44 &  70.92 ± 0.33 &  71.21 ± 0.35 &  70.19 ± 0.42 &  73.06 ± 0.16 &  72.29 ± 0.26 \\
4 & 0 & 0   & 0 &  75.71 ± 1.52 &  71.21 ± 0.26 &  71.28 ± 0.48 &  70.31 ± 0.34 &  72.25 ± 0.32 &  72.34 ± 0.35 \\
  & 1 & 1   & 0 &  74.03 ± 0.32 &  66.51 ± 1.64 &  64.79 ± 1.88 &  41.74 ± 2.26 &  21.09 ± 7.56 &  71.98 ± 0.45 \\
  &   &     & 1 &  76.28 ± 0.12 &  70.00 ± 0.26 &  70.60 ± 0.41 &  68.62 ± 0.36 &  72.16 ± 0.37 &  71.95 ± 0.34 \\
  &   & 20  & 0 &  75.41 ± 0.42 &  70.45 ± 0.19 &  70.32 ± 0.59 &  68.54 ± 0.25 &  69.24 ± 3.97 &  71.97 ± 0.35 \\
  &   &     & 1 &  76.57 ± 0.27 &  70.11 ± 0.22 &  71.01 ± 0.21 &  69.46 ± 0.36 &  71.64 ± 0.29 &  72.11 ± 0.34 \\
  &   & 50  & 0 &  75.01 ± 2.00 &  70.70 ± 0.43 &  70.42 ± 0.16 &  69.40 ± 0.49 &  72.59 ± 0.18 &  72.02 ± 0.33 \\
  &   &     & 1 &  75.59 ± 0.91 &  70.40 ± 0.19 &  71.25 ± 0.26 &  69.92 ± 0.38 &  72.47 ± 0.24 &  72.18 ± 0.07 \\
  &   & 100 & 0 &  76.20 ± 0.18 &  70.42 ± 0.34 &  70.89 ± 0.47 &  69.50 ± 0.36 &  72.98 ± 0.18 &  72.31 ± 0.33 \\
  &   &     & 1 &  75.63 ± 0.81 &  70.46 ± 0.40 &  71.20 ± 0.51 &  70.35 ± 0.17 &  72.92 ± 0.41 &  72.42 ± 0.24 \\
8 & 0 & 0   & 0 &  76.06 ± 0.25 &  70.69 ± 0.31 &  71.27 ± 0.27 &  70.21 ± 0.19 &  72.41 ± 0.22 &  72.57 ± 0.21 \\
  & 1 & 1   & 0 &  74.20 ± 0.33 &  67.82 ± 1.40 &  64.04 ± 1.67 &  44.47 ± 3.51 &  22.50 ± 3.62 &  72.14 ± 0.50 \\
  &   &     & 1 &  75.49 ± 0.49 &  70.20 ± 0.15 &  70.29 ± 0.41 &  68.81 ± 0.43 &  71.88 ± 0.24 &  71.71 ± 0.31 \\
  &   & 20  & 0 &  75.03 ± 0.62 &  70.58 ± 0.15 &  70.80 ± 0.44 &  68.58 ± 0.56 &  67.50 ± 1.86 &  71.99 ± 0.10 \\
  &   &     & 1 &  75.41 ± 1.26 &  70.29 ± 0.28 &  70.61 ± 0.25 &  69.15 ± 0.16 &  69.37 ± 1.30 &  72.03 ± 0.24 \\
  &   & 50  & 0 &  75.48 ± 0.93 &  70.59 ± 0.21 &  70.55 ± 0.44 &  69.59 ± 0.26 &  72.87 ± 0.30 &  72.34 ± 0.26 \\
  &   &     & 1 &  75.75 ± 0.61 &  70.34 ± 0.33 &  70.60 ± 0.65 &  69.33 ± 0.30 &  67.87 ± 1.50 &  72.71 ± 0.42 \\
  &   & 100 & 0 &  76.40 ± 0.31 &  70.65 ± 0.18 &  70.37 ± 0.40 &  69.60 ± 0.26 &  72.84 ± 0.12 &  72.62 ± 0.37 \\
  &   &     & 1 &  75.47 ± 1.37 &  70.43 ± 0.39 &  70.93 ± 0.51 &  69.96 ± 0.41 &  67.91 ± 2.05 &  72.22 ± 0.30 \\
\bottomrule
\end{tabular}
\end{sc}
\end{small}
\end{center}
\vskip -0.1in
\end{table}

\begin{table}[H]
\caption{\textbf{Test Accuracy on CIFAR-10 over 190 Epochs} It represents the mean accuracy and standard deviation for five seeds across each of the five models. The content in parentheses indicates the number of last layers used for the reset; if not specified, one layer is assumed. The meanings of each abbreviation in the table's headline are as follows: KL refers to self-distillation. K represents the number of teachers, where a value of zero indicates that self-distillation is not utilized. Regarding the reset, R indicates whether it is used; thus, when R is zero, reset is not utilized. $T_{\text{cycle}}$ signifies the duration of the cycle used for the teacher update and the student reset, expressed in epochs. M indicates the re-initialization method used for reset, where a value of one means that re-initialization was done using the mean of the teachers' weights. A value of zero signifies that random weights were utilized for re-initialization.}
\label{tbl:cifar10_190}
\vskip 0.15in
\begin{center}
\begin{small}
\begin{sc}
\begin{tabular}{|c|c|c|c|c|c|c|c|c|c|}
\toprule
 \multicolumn{1}{|c|}{KL} & \multicolumn{3}{|c|}{RESET} & \multicolumn{6}{|c|}{Model} \\
\cmidrule(lr){1-1} \cmidrule(lr){2-4} \cmidrule(lr){5-10}
 K & R & $T_{\text{cycle}}$ & M & googlenet  & mobilenetv2 & shufflenet & squeezenet & vgg(3) & vgg(1) \\
\midrule
0 & 0 & 0   & 0 &  94.58 ± 0.07 &  91.73 ± 0.26 &  92.33 ± 0.33 &  92.36 ± 0.11 &  93.62 ± 0.09 &  93.56 ± 0.17 \\
  & 1 & 1   & 0 &  94.27 ± 0.12 &  91.64 ± 0.14 &  91.71 ± 0.19 &  91.72 ± 0.41 &  93.46 ± 0.27 &  93.64 ± 0.22 \\
  &   &     & 1 &  94.56 ± 0.10 &  91.79 ± 0.18 &  92.33 ± 0.29 &  92.29 ± 0.19 &  93.60 ± 0.23 &  93.64 ± 0.22 \\
  &   & 20  & 0 &  94.29 ± 0.16 &  91.84 ± 0.15 &  92.11 ± 0.26 &  92.39 ± 0.18 &  93.61 ± 0.15 &  93.53 ± 0.05 \\
  &   &     & 1 &  94.62 ± 0.12 &  91.80 ± 0.19 &  92.27 ± 0.19 &  92.14 ± 0.22 &  93.68 ± 0.20 &  93.53 ± 0.05 \\
  &   & 50  & 0 &  94.58 ± 0.23 &  91.85 ± 0.26 &  92.44 ± 0.15 &  92.57 ± 0.31 &  93.56 ± 0.19 &  93.55 ± 0.13 \\
  &   &     & 1 &  94.64 ± 0.08 &  91.87 ± 0.20 &  92.51 ± 0.46 &  92.18 ± 0.18 &  93.73 ± 0.29 &  93.55 ± 0.13 \\
  &   & 100 & 0 &  94.60 ± 0.17 &  91.75 ± 0.25 &  92.40 ± 0.21 &  92.31 ± 0.11 &  93.73 ± 0.16 &  93.56 ± 0.19 \\
  &   &     & 1 &  94.65 ± 0.17 &  91.83 ± 0.06 &  92.26 ± 0.16 &  92.25 ± 0.14 &  93.67 ± 0.17 &  93.56 ± 0.19 \\
1 & 0 & 0   & 0 &  94.33 ± 0.19 &  91.96 ± 0.17 &  92.63 ± 0.21 &  92.35 ± 0.12 &  93.63 ± 0.13 &  93.57 ± 0.16 \\
  & 1 & 1   & 0 &  94.39 ± 0.32 &  91.59 ± 0.19 &  91.96 ± 0.13 &  91.43 ± 0.48 &  93.47 ± 0.16 &  93.68 ± 0.16 \\
  &   &     & 1 &  94.57 ± 0.17 &  91.72 ± 0.19 &  92.31 ± 0.17 &  92.59 ± 0.18 &  93.53 ± 0.11 &  93.68 ± 0.16 \\
  &   & 20  & 0 &  94.33 ± 0.16 &  91.69 ± 0.27 &  92.27 ± 0.20 &  92.23 ± 0.09 &  93.75 ± 0.14 &  93.62 ± 0.16 \\
  &   &     & 1 &  94.65 ± 0.15 &  91.76 ± 0.18 &  92.29 ± 0.31 &  92.29 ± 0.16 &  93.71 ± 0.06 &  93.63 ± 0.16 \\
  &   & 50  & 0 &  94.35 ± 0.18 &  91.84 ± 0.19 &  92.41 ± 0.20 &  92.31 ± 0.24 &  93.58 ± 0.16 &  93.71 ± 0.10 \\
  &   &     & 1 &  94.17 ± 0.35 &  91.76 ± 0.29 &  92.17 ± 0.27 &  92.55 ± 0.28 &  93.76 ± 0.19 &  93.71 ± 0.10 \\
  &   & 100 & 0 &  94.50 ± 0.12 &  91.73 ± 0.33 &  92.65 ± 0.17 &  92.18 ± 0.17 &  93.69 ± 0.14 &  93.67 ± 0.06 \\
  &   &     & 1 &  94.53 ± 0.20 &  92.02 ± 0.08 &  92.43 ± 0.32 &  92.44 ± 0.09 &  93.70 ± 0.13 &  93.67 ± 0.06 \\
2 & 0 & 0   & 0 &  94.48 ± 0.14 &  91.91 ± 0.13 &  92.41 ± 0.13 &  92.25 ± 0.21 &  93.60 ± 0.11 &  93.47 ± 0.17 \\
  & 1 & 1   & 0 &  94.34 ± 0.15 &  91.62 ± 0.14 &  91.74 ± 0.15 &  91.64 ± 0.27 &  93.32 ± 0.19 &  93.68 ± 0.06 \\
  &   &     & 1 &  94.47 ± 0.21 &  91.74 ± 0.17 &  92.09 ± 0.27 &  92.59 ± 0.11 &  93.65 ± 0.18 &  93.64 ± 0.17 \\
  &   & 20  & 0 &  94.24 ± 0.47 &  91.71 ± 0.18 &  92.44 ± 0.17 &  92.33 ± 0.16 &  93.62 ± 0.13 &  93.47 ± 0.10 \\
  &   &     & 1 &  94.67 ± 0.09 &  91.89 ± 0.27 &  92.22 ± 0.12 &  92.37 ± 0.22 &  93.66 ± 0.20 &  93.63 ± 0.21 \\
  &   & 50  & 0 &  94.38 ± 0.07 &  91.75 ± 0.33 &  92.35 ± 0.28 &  92.37 ± 0.28 &  93.61 ± 0.12 &  93.54 ± 0.11 \\
  &   &     & 1 &  94.29 ± 0.14 &  92.04 ± 0.26 &  92.30 ± 0.35 &  92.43 ± 0.26 &  93.64 ± 0.17 &  93.55 ± 0.17 \\
  &   & 100 & 0 &  94.30 ± 0.34 &  91.87 ± 0.31 &  92.60 ± 0.10 &  92.31 ± 0.26 &  93.65 ± 0.12 &  93.53 ± 0.19 \\
  &   &     & 1 &  94.53 ± 0.15 &  91.89 ± 0.26 &  92.33 ± 0.32 &  92.45 ± 0.17 &  93.74 ± 0.14 &  93.52 ± 0.07 \\
4 & 0 & 0   & 0 &  94.46 ± 0.11 &  91.87 ± 0.14 &  92.36 ± 0.18 &  92.44 ± 0.20 &  93.57 ± 0.11 &  93.65 ± 0.25 \\
  & 1 & 1   & 0 &  94.39 ± 0.05 &  91.86 ± 0.15 &  92.04 ± 0.08 &  91.85 ± 0.13 &  93.35 ± 0.18 &  93.61 ± 0.12 \\
  &   &     & 1 &  94.45 ± 0.23 &  91.93 ± 0.16 &  91.96 ± 0.16 &  92.33 ± 0.18 &  93.53 ± 0.14 &  93.48 ± 0.15 \\
  &   & 20  & 0 &  94.33 ± 0.09 &  91.76 ± 0.34 &  92.35 ± 0.29 &  92.35 ± 0.22 &  93.75 ± 0.36 &  93.71 ± 0.08 \\
  &   &     & 1 &  94.45 ± 0.20 &  91.86 ± 0.15 &  92.17 ± 0.19 &  92.67 ± 0.12 &  93.45 ± 0.20 &  93.57 ± 0.18 \\
  &   & 50  & 0 &  94.40 ± 0.22 &  91.92 ± 0.14 &  92.33 ± 0.22 &  92.45 ± 0.09 &  93.65 ± 0.21 &  93.65 ± 0.16 \\
  &   &     & 1 &  94.48 ± 0.21 &  91.84 ± 0.15 &  92.38 ± 0.12 &  92.47 ± 0.12 &  93.58 ± 0.12 &  93.72 ± 0.10 \\
  &   & 100 & 0 &  94.42 ± 0.14 &  91.74 ± 0.30 &  92.48 ± 0.23 &  92.44 ± 0.18 &  93.62 ± 0.18 &  93.65 ± 0.15 \\
  &   &     & 1 &  94.51 ± 0.23 &  91.84 ± 0.26 &  92.64 ± 0.15 &  92.45 ± 0.10 &  93.79 ± 0.07 &  93.57 ± 0.29 \\
8 & 0 & 0   & 0 &  94.60 ± 0.15 &  92.03 ± 0.22 &  92.54 ± 0.17 &  92.45 ± 0.19 &  93.56 ± 0.15 &  93.59 ± 0.07 \\
  & 1 & 1   & 0 &  94.39 ± 0.09 &  91.56 ± 0.25 &  92.08 ± 0.12 &  91.56 ± 0.30 &  93.32 ± 0.24 &  93.70 ± 0.11 \\
  &   &     & 1 &  94.31 ± 0.32 &  91.90 ± 0.16 &  91.63 ± 0.19 &  92.35 ± 0.24 &  93.65 ± 0.19 &  93.74 ± 0.13 \\
  &   & 20  & 0 &  94.28 ± 0.42 &  91.83 ± 0.22 &  92.46 ± 0.27 &  92.45 ± 0.28 &  93.68 ± 0.12 &  93.70 ± 0.05 \\
  &   &     & 1 &  94.47 ± 0.14 &  91.80 ± 0.21 &  92.24 ± 0.22 &  92.55 ± 0.33 &  93.15 ± 0.27 &  93.76 ± 0.10 \\
  &   & 50  & 0 &  94.38 ± 0.18 &  91.95 ± 0.25 &  92.39 ± 0.16 &  92.34 ± 0.26 &  93.66 ± 0.10 &  93.71 ± 0.18 \\
  &   &     & 1 &  94.48 ± 0.18 &  91.77 ± 0.15 &  92.35 ± 0.16 &  92.47 ± 0.30 &  93.11 ± 0.49 &  93.71 ± 0.20 \\
  &   & 100 & 0 &  94.34 ± 0.18 &  91.66 ± 0.27 &  92.52 ± 0.11 &  92.22 ± 0.17 &  93.64 ± 0.20 &  93.58 ± 0.07 \\
  &   &     & 1 &  94.47 ± 0.16 &  91.80 ± 0.16 &  92.51 ± 0.24 &  92.41 ± 0.17 &  93.14 ± 0.17 &  93.65 ± 0.12 \\
\bottomrule
\end{tabular}
\end{sc}
\end{small}
\end{center}
\vskip -0.1in
\end{table}


\end{document}